\documentclass[journal]{IEEEtran}
\usepackage{float}
\usepackage{graphicx}
\usepackage{caption}
\usepackage{amsmath}
\usepackage{amssymb}
\usepackage{algorithm}
\usepackage{booktabs}
\usepackage{multirow}
\usepackage{xcolor}  % F
\usepackage{enumitem}
\usepackage{graphicx}   % For including graphics/images
\usepackage{subfig}     % For subfigures using \subfloat
\usepackage{caption}    % For caption 
\usepackage{algorithmic}
\title{ProDAT: Progressive Density-Aware Tail-Drop for Point Cloud Coding}
\author{Zhe~Luo\thanks{Z. Luo, W. Jia, and S. Perry are with the School of Electrical and Data Engineering, University of Technology Sydney, Sydney, NSW 2007, Australia (e-mail: zhe.luo-1@student.uts.edu.au).},
        Stuart~Perry,
        Wenjing~Jia}
\begin{document}
\maketitle

\begin{abstract}
Three-dimensional (3D) point clouds are becoming increasingly vital in applications such as autonomous driving, augmented reality, and immersive communication, demanding real-time processing and low latency. However, their large data volumes and bandwidth constraints hinder the deployment of high-quality services in resource-limited environments. 
Progressive coding, which allows for decoding at varying levels of detail, provides an alternative by allowing initial partial decoding with subsequent refinement. 
Although recent learning-based point cloud geometry coding methods have achieved notable success, their fixed latent representation does not support progressive decoding. 
%This limitation arises from their design, which encodes the entire point cloud into a fixed latent representation. % without a hierarchical structure. 
To bridge this gap, we propose ProDAT, a novel density-aware tail-drop mechanism for progressive point cloud coding. By leveraging density information as a guidance signal, latent features and coordinates are decoded adaptively based on their significance, therefore achieving progressive decoding at multiple bitrates using one single model. 
%This allows us to achieve progressive decoding at multiple bitrates using one single model, while preserving critical geometric and structural details at various stages. 
%
Experimental results on benchmark datasets show that the proposed ProDAT not only enables progressive coding but also achieves superior coding efficiency compared to state-of-the-art learning-based coding techniques, with over 28.6\% BD-rate improvement for PSNR-D2 on SemanticKITTI and over 18.15\% for ShapeNet. 
%These results confirm that our method provides superior coding efficiency and meets the demands of progressive geometry coding for point clouds.
\end{abstract}

\begin{IEEEkeywords}
Progressive coding, %Point cloud coding, Point cloud coding, 
Point cloud compression, %Deep learning, 
Tail-drop, Density-aware.
\end{IEEEkeywords}

\section{Introduction}
 
Point clouds, generated by 3D capturing technologies such as Light Detection and Ranging (LiDAR) scanners, provide detailed 3D representations of environments via millions of spatial points with attributes like color, intensity, and reflectivity. 
Unlike 2D images with regular pixel grids, point clouds are inherently irregular and unstructured, presenting significant challenges for storage, transmission, and processing~\cite{quach2022survey}. 
A single LiDAR scan can generate hundreds of millions of points, resulting in substantial storage and bandwidth demands~\cite{cao2025real,golla2015real}. Consequently, point cloud coding (PCC) is essential for practical applications, including autonomous driving, virtual and augmented reality (VR/AR), and 3D mapping~\cite{liu2024att,yue2018lidar,fukuda2018point}. 
However, the irregular structure of point clouds makes conventional 2D and video coding methods unsuitable, necessitating specialized PCC techniques to preserve geometric fidelity efficiently.

Traditional PCC methods, such as Geometry-based point cloud compression (G-PCC) and Video-based Point Cloud Compression (V-PCC)~\cite{mdgc20207,mdgc20208}, encode geometry hierarchically or project 3D data onto 2D planes to exploit video codecs, achieving widespread adoption~\cite{quach2022survey}. However, they falter on large-scale datasets, such as SemanticKITTI~\cite{behley2019semantickitti}, exhibiting low coding efficiency, degraded reconstruction quality, and high computational overhead. 
In contrast, learning-based PCC leverages deep learning to better capture spatial and structural complexities~\cite{guarda2022ipleiria,wang2021lossy,wang2022sparse}, drawing from 2D innovations like hyperprior variational models~\cite{balle2018variational} to outperform traditional codecs at low bitrates~\cite{guarda2022ipleiria}. 
Nevertheless, they typically produce a single, monolithic bitstream that requires full decoding for reconstruction, which is inadequate for time-sensitive applications such as autonomous driving. Thus, progressive coding represents a pivotal advancement for large-scale point cloud processing, blending practical benefits with theoretical novelty. 

In 2D image and video domains, progressive coding is a well-established technique for efficient data delivery and real-time adaptability~\cite{lee2022dpict}. For images, the JPEG standard~\cite{wallace1992jpeg} uses multi-scan encoding for incremental detail enhancement, aiding previews on bandwidth-limited networks via initial low-quality representations refined progressively. Similarly, JPEG 2000~\cite{taubman2002jpeg2000} leverages wavelet transforms to provide scalable resolution and quality, supporting progressive decoding that adapts flexibly to user needs or network conditions. 
In the video domain, the H.264 Scalable Video Coding (SVC) extension~\cite{wiegand2003overview} adopts a layered architecture: a base layer provides foundational quality, with enhancement layers incrementally improving resolution, frame rate, or fidelity to enable adaptive streaming amid varying network conditions or device capabilities. 
Together, these standards exemplify progressive coding's advantages—reduced latency and optimized resource utilization—as foundations of modern multimedia systems.

\begin{figure*}[ht]
    \centering
    \includegraphics[width=1\textwidth]{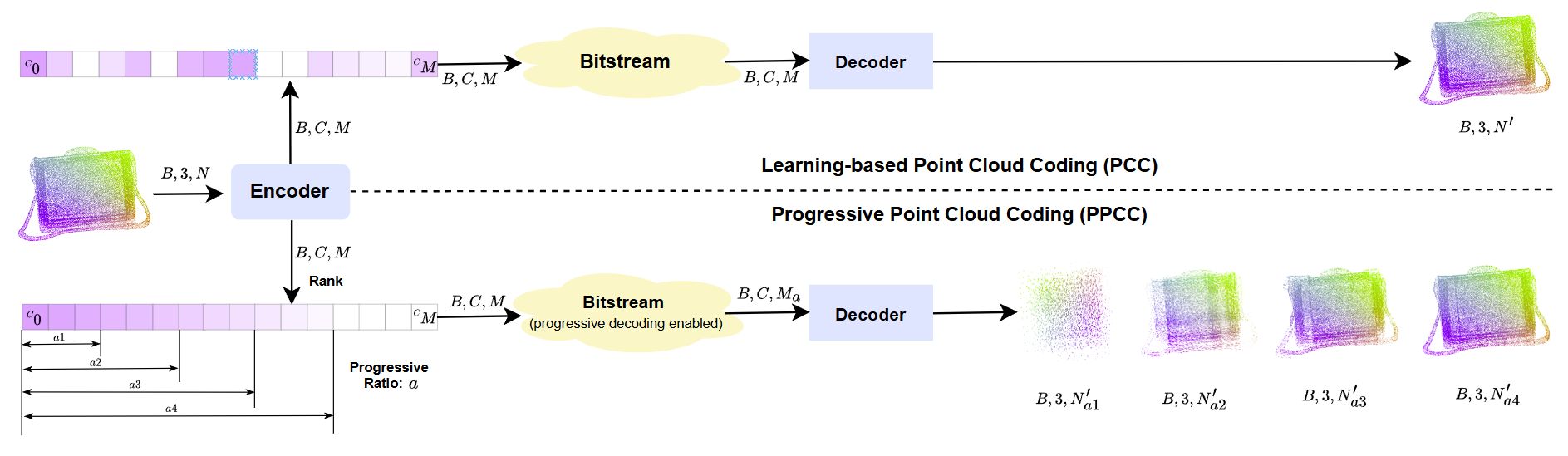}
    
    % \caption{Comparisons between Point Cloud Coding (PCC) and Progressive Point Cloud Coding (PPCC). The primary distinction is that PPCC, utilizing ProDAT, generates a progressive bitstream post-encoding, enabling customized decoding for variable coding ratios $a$ and reconstruction qualities. This bitstream, derived from progressive training, allows users to set a customized Progressive Ratio (PR) $\alpha$ to achieve different bits per point (BPP) and decoding results ($N'_{a1}$ to $N'_{a4}$).}
    \caption{Comparisons between Point Cloud Coding (PCC) and Progressive Point Cloud Coding (PPCC). PPCC generates a progressive bitstream post-encoding, enabling customized decoding for variable coding ratios $a$ and reconstruction qualities.}    
    \label{fig:comparison}
    \vspace{-1em}
\end{figure*}

In the point cloud domain, progressive point cloud coding (PPCC) remains underexplored compared to its well-established counterparts in 2D images and video. 
Traditional approaches, such as Huang \textit{et al.}~\cite{huang2006octree}, estimate geometric centers of tree-front cells (\textit{i.e.}, nonempty cells at the current octree level) for progressive coding but are hindered by their computational complexity. 
Recent deep learning advancements, \textit{e.g.}, Rudolph \textit{et al.}~\cite{rudolph2024progressive}, leverage quantization residuals and entropy bottleneck transformations to enable progressive attribute coding. 
However, progressive geometry coding, particularly on large-scale benchmark datasets such as SemanticKITTI~\cite{behley2019semantickitti} and ShapeNet~\cite{wu20153d}, remains understudied, highlighting a critical research gap and the need for scalable, learning-based solutions for real-time applications. 

Meanwhile, conventional and learning-based PCC techniques typically use single-rate encoding, producing a fixed bitstream that %necessitates complete decoding 
must be fully decoded to attain maximum fidelity. 
This process is depicted in Fig.~\ref{fig:comparison}, where the input point cloud comprises \(N\) points, and the reconstructed point cloud contains \(N'\) points. 
While effective for static storage, single-rate encoding %proves 
is suboptimal in dynamic, bandwidth-constrained environments %requiring 
that demand low-latency access and incremental refinement. %, where the inability to adaptively decode bitstream portions hampers practicality. 
The inability to decode partial bitstreams limits its practicality.
%Conversely, 
By contrast, PPCC enables flexible partial decoding, providing an initial coarse representation that progressively refines. %with added data. 
%For example, as also demonstrated 
As illustrated in Fig.~\ref{fig:comparison}, progressive point cloud coding results are obtained during testing by varying the Progressive Ratio (PR) $\alpha$. Each $\alpha$ corresponds to a specific bit-per-point (BPP) and rate-distortion value, %which scale systematically 
scaling from low to high %from $N'_{a1}$ to $N'_{a4}$ (we choose four PR values here), 
across four PR values, and yielding reconstruction $N'_{a1}$ to $N'_{a4}$. %allowing for 
This enables a controlled trade-off between data rate and reconstruction quality.

Inspired by prior work~\cite{koike2020stochastic,hojjat2023progdtd}, we introduce a \textbf{d}ensity-\textbf{a}ware \textbf{t}ail-drop progressive point cloud coding (ProDAT) method. 
Similar to Progressive JPEG~\cite{wallace1992jpeg} and H.264 Scalable Video Coding (SVC)~\cite{wiegand2003overview}, ProDAT enables %facilitates 
incremental decoding by training the model to selectively discard less critical %latent 
features, enabling progressive quality improvements at higher bitrates during decoding. 
Unlike 2D and video counterparts %reliant on 
that rely on uniform grids or temporal continuity, %our method confronts 
our approach addresses the irregular, unstructured geometry of point clouds by explicitly leveraging density. 
%Density, a pivotal factor in point cloud processing~\cite{schwarz2018emerging}~\cite{he2022density}, serves as an indicator of local detail complexity. 
High-density regions %often signify 
typically indicate intricate geometry (\textit{e.g.}, object surfaces), whereas sparse areas %typically denote 
usually correspond to simpler, less critical structures. %spaces. 
By exploiting density variations, ProDAT prioritizes structurally significant details, %enhancing 
improving rate-distortion efficiency through %smarter 
informed feature selection. In detail, our approach delivers the following key contributions:

\begin{itemize}
    \item %\textbf{Novel Progressive coding Method:} 
    %We propose a solution named ProDAT, which enables flexible progressive point cloud coding regarding benchmark datasets.
    We propose ProDAT, a novel framework that achieves progressive point cloud coding with a single training stage. % iteration. %, significantly reducing computational overhead while maintaining competitive performance.
    \item %\textbf{Density-Aware Tail-Drop Module:} 
    We develop a density-aware tail-drop %data selection 
    approach that leverages point and structure density to prioritize structurally significant regions, enhancing efficiency by focusing on complex, high-density areas. % while adaptively compressing sparse ones.
    % \item %\textbf{Single-Training, Multi-Bitrate Capability:} 
    % Through a single training process, ProDAT supports multiple bitrates, offering flexibility and eliminating retraining needs.
    \item %\textbf{Superior coding Performance:} 
    We validate ProDAT on SemanticKITTI and ShapeNet, demonstrating superior BD-Rate performance compared to state-of-the-art methods.
    % Evaluations on SemanticKITTI and ShapeNet show ProDAT outperforms leading learning-based methods, achieving BD-rate gains of over 28.6\% on SemanticKITTI and 18.15\% on ShapeNet for high-quality, low-bitrate outputs.
\end{itemize}

% In summary, ProDAT establishes a new benchmark in progressive point cloud coding by integrating density-aware tail-drop with progressive refinement, delivering efficiency and adaptability for advanced 3D data streaming applications.
The remainder of this paper is organized as follows: Section~\ref{sect:review} provides a concise overview of existing point cloud coding and related image progressive coding strategies. 
Section~\ref{sect:method} details the proposed ProDAT framework, including the Density-aware Tail-drop strategy. 
Experimental results with ablation studies are presented in Section~\ref{sect:exp}, followed by a discussion and conclusion in Section~\ref{sect:conclusion}.

\section{Related Work}
\label {sect:review}

\subsection{Point Cloud Coding}

Traditional point cloud coding techniques employ spatial data structures and quantization to reduce redundancy without neural networks. For example, MPEG's G-PCC~\cite{mdgc20207} uses octree representations and triangular surface coding for geometric encoding in static scenes, while V-PCC~\cite{mdgc20208} projects 3D data onto 2D planes to exploit video codecs for temporal redundancy in dynamic sequences. However, these methods incur high computational complexity, substantial storage needs, and limited scalability across datasets. To mitigate these limitations, learning-based methods leverage deep neural networks for compact, efficient representations, surpassing traditional approaches in scalability and performance.

Point-based methods process raw points directly, avoiding losses from intermediate transformations. Inspired by PointNet++~\cite{qi2017pointnet++}, Hao \textit{et al.}’s KNN-based encoding~\cite{xu2025fast} and D-PCC’s density preservation~\cite{he2022density} enhance reconstruction quality and adaptability across sparse or irregular distributions. Octree-based methods refine hierarchical tree coding with neural networks. OctSqueeze~\cite{huang2020octsqueeze} introduces probabilistic occupancy modeling, followed by Tingyu \textit{et al.}’s hierarchical latent variables~\cite{fan2023multiscale}. 
Attention mechanisms in OctAttention~\cite{fu2022octattention} further optimize context modeling, balancing efficiency and detail preservation. Voxel-based methods treat point clouds as occupancy grids, applying 3D CNNs. Quach \textit{et al.}’s convolutional autoencoder~\cite{quach2019learning}, PCGC~\cite{wang2021lossy}, and JPEG Pleno~\cite{guarda2022ipleiria} pioneered this approach, while sparse convolutions~\cite{wang2022sparse} and block-based partitioning~\cite{guarda2019point} address sparsity inefficiencies. These learning-based strategies outperform traditional techniques, offering robust adaptability to diverse point cloud densities and structures with enhanced efficiency and reconstruction fidelity.

\subsection{Progressive Image Coding}

Progressive coding has been extensively explored in learning-based image coding, building on %drawing from 
foundational works by Ballé \textit{et al.}~\cite{balle2016end,balle2018variational} and Minnen \textit{et al.}~\cite{minnen2018joint}. 
Early efforts partitioned latent representations into a base layer and enhancement layers, each decoded by separate networks to produce intermediate image versions~\cite{cai2019novel}. 
Others employed Recurrent Neural Networks (RNNs) to progressively encode quantization residuals~\cite{diao2020drasic,islam2021image,johnston2018improved}, but these approaches suffer from high computational costs, memory usage, and low throughput~\cite{huang2024unveiling}.

Alternatively, Lu \textit{et al.}~\cite{lu2021progressive} proposed nested quantization, defining multiple levels with nested grids to progressively refine all latents from coarsest to finest. 
Building on this, Lee \textit{et al.}~\cite{lee2022dpict} represented encoded features in ternary digits (trits) and transmitted them %, encoding them 
in decreasing significance order using rate-distortion priorities, % to transmit 
sending critical information first. 
However, this results in suboptimal performance and degraded quality at low bitrates, often requiring a post-processing network for refinement. 
Similarly, Li \textit{et al.}~\cite{li2022learned} replaced uniform quantizers with dead-zone quantizers %in their learned progressive coding framework to improve the nested quantization. This substitution 
to reduce redundant symbols, thereby improving the efficiency of progressive coding. 
Despite strong progressive coding performance in images, these methods~\cite{lee2022dpict,li2022learned} rely on fixed-rate models, limiting their flexibility.

To overcome the limitations of RNN and nested quantization, tail-drop techniques enable variable-rate coding, reducing computational complexity while improving coding efficiency. 
Koike \textit{et al.}~\cite{koike2020stochastic} showed that discarding the least important principal components %can be discarded to realize 
yields variable rate dimensionality reduction with graceful degradation. %that gracefully degrades the distortion. % By utilizing a variable-rate coding model, dropping the unimportant channels in the latent features can be an effective way to do progressive coding. 
Based on this, Hojjat \textit{et al.}~\cite{hojjat2023progdtd} introduced a double-tail-drop progressive training protocol that prioritizes latent and hyper-latent channels by %based on 
relevance~\cite{balle2016end,balle2018variational}, % to do the progressive coding for images, including both latent features and the hyper-latent features~\cite{balle2016end,balle2018variational}. This approach allows users 
%to transmit data according to their importance in the order of their sorted index, eliminating the need for nested quantization.
enabling transmission in order of channel importance and eliminating the need for nested quantization.

\subsection{Progressive Point Cloud Coding}

Although progressive coding is well established in 2D images, its application to point clouds remains underexplored. 
%Nonetheless, 
Notable advancements include %traditional methods, such as 
Huang \textit{et al.}'s~\cite{huang2006octree} octree-based subdivision, which estimates geometric centers of tree-front cells to support progressive coding. 
%They also utilize a generic encoder that exploits attribute-specific characteristics at each subdivision level~\cite{huang2008generic}.

\begin{figure*}[ht]
    \centering
    \includegraphics[width=1.0\textwidth]{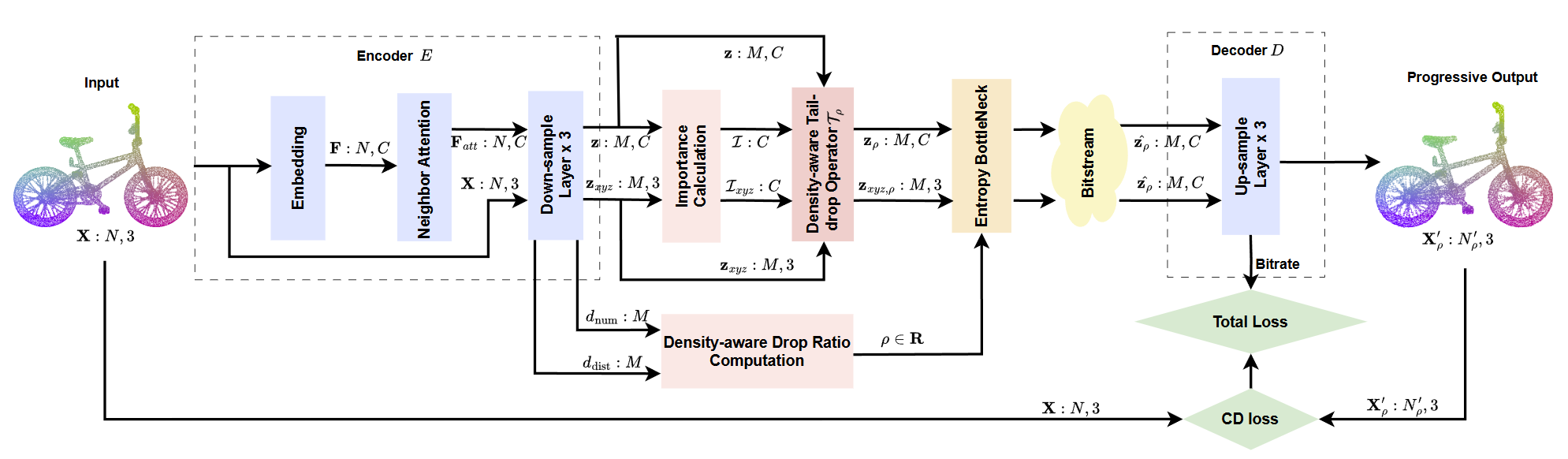}
    \captionsetup{width=1.0\textwidth} 
    % \caption{This figure illustrates the architecture of ProDAT. Our progressive coding model for point clouds leverages an end-to-end autoencoder with specialized blocks, including an Encoder $E$, a Density-aware Tail-drop Operator, an Entropy Bottleneck, and a Decoder $D$. The total loss is formed by a trade-off between density loss, Chamfer Distance (CD) loss, and Bitrate. 
    % Further technical details will be described in Sect.~\ref{ProDAT_Framework}.}
    \caption{The architecture of the proposed ProDAT. The progressive coding model of ProDAT leverages an end-to-end autoencoder which consists of an Encoder $E$, a Density-aware Tail-drop Operator, an Entropy Bottleneck, and a Decoder $D$. The total loss is defined as a trade-off among density loss, Chamfer Distance (CD) loss, and bitrate. See Sect.~\ref{ProDAT_Framework} for further details.}
    \label{fig:DATD_architecture}
    \vspace{-1em}
\end{figure*}

%Researchers are using 
Recent work applies deep learning to do progressive point cloud coding. 
Rudolph \textit{et al.}~\cite{rudolph2024progressive} employed quantization residuals from prior representations with a learned lightweight transformation in the entropy bottleneck to enable progressive attribute coding. 
Similarly, Gokulnath \textit{et al.}~\cite{vadivel2021progressive} %preprocess the colored point cloud into a sequence of view-specific six-dimensional (6D) images by projecting the 3D model from multiple viewpoints, where each image encodes RGB color and XYZ geometry in a 2D grid, followed by applying a symmetry-based convolutional neural pyramid to progressively encode these images from coarse to fine levels, exploiting redundancies in the projections for efficient transmission. 
projected the 3D model from multiple viewpoints to form a sequence of view-specific six-dimensional (RGB+XYZ) images on 2D grids, then used a symmetry-based convolutional neural pyramid to encode them progressively from coarse to fine, exploiting inter-projection redundancies for efficient transmission. 
Nonetheless, progressive geometry coding for point clouds remains underexplored, particularly on challenging benchmarks like SemanticKITTI~\cite{behley2019semantickitti} and ShapeNet~\cite{wu20153d}. 
The vast scale of LiDAR data and the geometric diversity of models pose significant challenges for PCC in both data volume and detail preservation. 
%These challenges draw the significance of progressive PCC lies in the capability to enhance data efficiency and real-time applicability, particularly in resource-intensive fields like autonomous navigation and immersive media~\cite{rudolph2024progressive,vadivel2021progressive}. 
These challenges underscore the significance of progressive PCC for improving data efficiency and real-time applicability in resource-intensive domains such as autonomous navigation and immersive media~\cite{rudolph2024progressive,vadivel2021progressive}. 
To address this need, we propose ProDAT, which leverages %utilizes 
density information to guide progressive coding and prioritize critical regions of the point cloud. 
Through tail-drop, it selectively preserves essential features, %alleviating decompression 
reducing decoder computational load while enabling %facilitating 
efficient, controllable progressive decoding.

\section{Methodology}
\label {sect:method}

% Our ProDAT architecture implements a comprehensive point cloud coding framework through an integrated pipeline of specialized modules. The process starts with an input block that feeds point cloud data into an embedding module, which transforms raw coordinates into high-dimensional feature representations via a lightweight pre-convolution network. Subsequently, a multi-hierarchical encoder employs Furthest Point Sampling (FPS) and local attention mechanisms to downsample the point cloud while preserving critical geometric information and encoding density statistics at multiple scales. A novel Density-aware Tail-drop operator then analyzes the latent space information based on local point distribution characteristics, selectively pruning channels while retaining the essential ones, which enables the progressive coding capabilities. These pruned representations are then quantized through an entropy bottleneck to compress the features and coordinates into a bitstream. The decoder, which comprises three upsampling layers, reconstructs the point cloud by predicting the dropped channels and refining geometry. The entire framework is optimized using a loss function that incorporates Chamfer distance and bitrate to ensure high-fidelity reconstruction at competitive bitrates with progressive performance.

\subsection{Problem Definition}

A point cloud can be represented as a matrix \(\mathbf{X} \in \mathbb{R}^{3 \times N}\), where each column \(\mathbf{x}_i \in \mathbb{R}^3\) denotes the coordinates of the \(i\)-th point, for \(i = 1, 2, \dots, N\). 
Learning-based point cloud coding aims to learn an effective neural coding model $\mathcal{F}_\theta$ that maps the input to a compressed and subsequently reconstructed form:
\begin{equation}
    \mathcal{F}_\theta: (\mathbf{X}) \mapsto \mathcal{B} \mapsto (\mathbf{X}'),
    \label{eq2}
\end{equation}
where $\mathbf{X}'$ is the reconstructed coordinates, and $\mathcal{B}$ represents the complete monolithic bitstream requiring full decoding for reconstruction in standard coding.

%In this work, 
We consider two coding paradigms. 
Traditional (non-progressive) learning-based point cloud coding optimizes a fixed rate-distortion trade-off by minimizing the loss function:
\begin{equation}
    \mathcal{L} = \mathcal{D}(\mathbf{X}, \mathbf{X}') + \lambda \mathcal{R},
    \label{eq1}
\end{equation}
where $ \mathcal{D}(\mathbf{X}, \mathbf{X}') $ is the distortion metric, commonly the Chamfer Distance~\cite{perry2020quality}, measuring the discrepancy between the original point cloud $\mathbf{X}$ and its reconstruction $\mathbf{X}'$; $\mathcal{R}$ %$\mathbf{R}$ 
represents the bit rate, typically derived from the entropy of quantized latent representations; and $\lambda$ is a parameter that controls the trade-off between distortion and bitrate.

In contrast, progressive point cloud coding (PPCC) structures latents for incremental reconstruction from a single bitstream. In ProDAT, we achieve the PPCC by ranking latent channels by variance-based importance, enabling rate-scalable decoding through selective channel activation while preserving rate-distortion efficiency across different bitrates. To better demonstrate the progressive coding's capability and performance, we introduce a controllable Progressive Ratio (PR) $\alpha \in [0,1]$ specifying the fraction of latent channels to activate in decompression, enabling rate-scalable reconstruction:
\begin{equation}
    \mathcal{F}_\theta^{\text{prog}}: (\mathbf{X}) \mapsto \mathcal{B} \mapsto \mathcal{B}_\alpha \mapsto (\mathbf{X}'_\alpha),
    \label{eq3}
\end{equation}
where $\mathcal{B}_\alpha \subset \mathcal{B}$ represents a subset of the total bitstream containing approximately $\alpha$ proportion of the total bits. In decompression, the reconstruction quality monotonically improves with $\alpha$, while computational cost and bitrate scale proportionally. This enables dynamic adaptation, allowing decoders to begin with minimal channels for coarse reconstruction and progressively refine quality by incorporating additional channels as bandwidth permits.

\subsection{The ProDAT Framework}
\label {ProDAT_Framework}

Our framework extends the density-preserving D-PCC architecture~\cite{he2022density} by transforming a monolithic coding model into a progressive one with density-aware adaptation capabilities. 
%As shown in Algorithm~\ref{alg:prodat}, 
% We introduce the red section to highlight our method's contribution to enabling efficient progressive coding of point clouds. 
As shown in Fig.~\ref{fig:DATD_architecture}, the overall design follows an autoencoder architecture %with specialized components 
specialized for point cloud processing and entropy-constrained coding. 
%our model architecture includes an encoder with 
The encoder consists of three downsampling stages, configurable with factors of 1/2, 1/3, or 1/4, and a decoder with a maximum upsampling factor of 8. 
% The training procedure for progressive point cloud coding is formalized as:
The progressive coding process is formalized as:
\begin{equation}
\begin{aligned}
    \mathbf{F} &= F(\mathbf{X}), \\
    \mathbf{z}, \mathbf{z}_{xyz}, \mathbf{d} &= E(\mathbf{X}_{xyz}, \mathbf{F}), \\
    {\mathbf{z}}_{\rho}, {\mathbf{z}}_{xyz,\rho} &= \mathcal{T}_{\rho}(\mathbf{z}, \mathbf{z}_{xyz}, \mathbf{d}), \\
    \hat{\mathbf{z}_{\rho}}, \hat{\mathbf{z}}_{xyz, \rho} &= \mathcal{B}_{z}({\mathbf{z}}_{\rho}), \mathcal{B}_{xyz}({\mathbf{z}}_{xyz,\rho}), \\
    R_{\rho} &= \text{BPP}(\hat{\mathbf{z}_{\rho}}, \hat{\mathbf{z}}_{xyz, \rho}), \\
    \mathbf{X}_{\rho}' &= D(\hat{\mathbf{z}_{\rho}}, \hat{\mathbf{z}}_{xyz, \rho}).
\label{eq:prodat_train}
\end{aligned}
\end{equation}
Here, $F$ extracts initial features $\mathbf{F} \in \mathbb{R}^{C \times N}$, which are refined by the encoder $E$ through a series of down-sampling layers equipped with a point transformer mechanism. 
This yields downsampled coordinates $\mathbf{z}_{xyz} \in \mathbb{R}^{3 \times M}$, latent features $\mathbf{z} \in \mathbb{R}^{C \times M}$, and density statistics $\mathbf{d} \in \mathbb{R}^M$. 
Here, \(M\) denotes the number of down-sampled points and \(C\) the feature channels. 
The density-aware tail-drop operator $\mathcal{T}{\rho}$ prunes latent representations according to a drop ratio $\rho \in [0,1]$ (linked to the Progressive Ratio $\alpha = 1 - \rho$). 
Quantization is performed by Entropy bottlenecks \(\mathcal{B}_{\mathbf{z}}\) and \(\mathcal{B}_{\mathbf{z}_{xyz}}\), from which the bitrate $R_{\rho}$ is computed. 
The decoder $D$ reconstructs $\mathbf{X}'_{\rho} \approx \mathbf{X}$, with reconstruction quality directly controlled by $\rho$.

The density-aware tail-drop mechanism is pivotal for enabling progressive coding. 
Training employs a stochastic drop ratio $\rho$ as in~\cite{koike2020stochastic,hojjat2023progdtd}, which is dynamically adjusted based on local density statistics~\cite{he2022density}. 
%Through this training paradigm, 
This allows the model to prioritize dense regions while adaptively handling sparse areas, making it robust to incomplete feature representations. %, effectively acquiring density-aware coding capabilities. 
% This stochastic training strategy enhances the model's robustness to partial feature representations, 
This stochastic training strategy allows the model to reconstruct point clouds from varying levels of latent feature completeness during progressive coding at inference based on customized \(\rho \in [0,1]\). 
%This testing framework eliminates the need for retraining at different coding rates, enabling efficient progressive point cloud coding through a single trained model.
At inference, varying $\rho$ produces reconstructions at different bitrates without retraining, enabling efficient progressive coding with a single trained model.

\subsection{Channel-Importance based Density-aware Tail Drop}
\label{density_define}
%\begin{figure*}[ht]
    %\centering
    %\includegraphics[width=1\textwidth]%{./figures/Networks/Comparison_different_drop.png}
    %\captionsetup{width=1.0\textwidth} 
    %Uniform channel dropping~\cite{hojjat2023progdtd, koike2020stochastic} applies a fixed dropping ratio across all geometries, irrespective of their structural variations. Our density-aware channel dropping operator employs adaptive dropping ratios (\textit{e.g.}, 1/8, 2/8, 3/8) tailored to local geometry density in corresponding high-, medium-, and low-density regions, respectively, achieving superior rate-distortion performance to uniform approaches. %This approach retains more channels in high-density regions to preserve critical details and reduces channels in low-density regions for efficiency. Consequently, our method achieves superior rate-distortion performance compared to uniform approaches.
    %}
    %\label{fig:channel_drop_difference}
    %\vspace{-1em}
%\end{figure*}

% \subsubsection{Density-Awareness Methodology}
\subsubsection{Density-Aware Tail Drop}

In contrast to uniform channel drop techniques in all regions, such as those implemented in ProgDTD~\cite{hojjat2023progdtd} and~\cite{koike2020stochastic}, our approach introduces a density-aware tail-drop mechanism with purpose of adapting to the local density characteristics of the point cloud.%: a lower drop ratio is applied in high-density regions to retain intricate details, whereas a higher drop ratio is applied in low-density regions to enhance coding efficiency. 
% This density-aware tail-drop approach optimizes channel allocation by preserving finer details in complex areas while allowing increased coding in flat areas for both feature and coordinate representations.

% The core innovation lies in our adaptive channel selection mechanism:
% \begin{equation}
% \rho = \rho_{\max} - (\rho_{\max} - \rho_{\min}) \cdot \delta,
% \label{fig:density_eq_1}
% \end{equation}
% where 
% \begin{equation}
% \delta= \frac{1}{2} \left( \frac{d_{\text{num}}}{d_{\max}} + \left(1 - \frac{d_{\text{dist}}}{m_{\max}}\right) \right).
% \label{fig:density_eq_2}
% \end{equation}
% Here, $\rho$ is the density-aware drop ratio, and $\delta$ is a composite density score derived from local densities. $\delta$ governs channel preservation during coding: regions with elevated $\delta$, reflecting higher geometric complexity, experience smaller drop ratios, retaining more channels to ensure information fidelity. In contrast, regions with lower $\delta$, denoting simpler geometry, undergo higher channel dropping. 

Specifically, for point cloud $\mathbf{P} = \{p_1, ..., p_N\}$ and its downsampled set $\mathbf{X} = \{x_1, ..., x_M\}$, 
the density-aware drop ratio, denoted by $\rho$, %, in our adaptive channel selection mechanism 
is defined as:
\begin{equation}
\rho = \rho_{\max} - (\rho_{\max} - \rho_{\min}) \cdot \delta,
\label{fig:density_eq_1}
\end{equation}
where $\rho_{\min}$ and $\rho_{\max}$ %are the boundary parameters that 
define the range of the drop ratio, empirically set to $0.15$ and $0.4$, and $\delta$ is the composite density score, governing channel preservation during coding, which is %derived from local densities as: 
calculated by averaging normalized point concentration with inverted normalized distance as: 
\begin{equation}
\delta= \frac{1}{2} \left( \frac{d_{\text{num}}}{d_{\max}} + \left(1 - \frac{d_{\text{dist}}}{m_{\max}}\right) \right).
\label{fig:density_eq_2}
\end{equation}
Here, $d_{\text{num}}$ quantifies local point concentration and $d_{\text{dist}}$ captures spatial distribution; 
$d_{\max}$ and $m_{\max}$ are their expected upper bounds, dynamically updated during training to normalize density metrics to the range of $[0,1]$ across different datasets. 
For each downsampled point $x_i$, $d_{\text{num}}$ is defined by counting points near $p_i$ that collapse to $x_i$ as:
% Thus, regions with a higher density score $\delta$, indicating greater geometric complexity, have smaller drop ratios and retain more channels to preserve information, whereas regions with lower $\delta$, representing simpler geometry, have more channels dropped. 
%Empirically, we set $\rho_{\min} = 0.15$ and $\rho_{\max} = 0.4$ here, based on extensive experiments, to enhance the model's ability to handle dropped channels and achieve strong progressive coding performance. 
%
% Derived from D-PCC~\cite{he2022density}, $d_{\text{num}}$ quantifies local point concentration by counting points collapsing to each downsampled location, and $d_{\text{dist}}$ captures spatial distribution by measuring the distance between original points and their downsampled representative. 
%
% Specifically, for each downsampled point $x_i$ from the downsampled set $\mathbf{X} = \{x_1, ..., x_M\}$ and original point cloud $\mathbf{P} = \{p_1, ..., p_N\}$:
\begin{equation}
d_{\text{num}}(x_i) = |\{p_j : \text{NN}(p_j) = x_i\}|,
\label{eq:d_num}
\end{equation}
and $d_{\text{dist}}$ measures the total distance between $x_i$ and each original point in the neighborhood of $p_i$ that collapses to $x_i$:
\begin{equation}
d_{\text{dist}}(x_i) = \frac{1}{d_{\text{num}}(x_i)} \sum_{p_j \in \text{NN}^{-1}(x_i)} \|p_j - x_i\|_2.
\label{eq:d_dist}
\end{equation}
Here, $\text{NN}(p_j)$ returns the nearest downsampled point to $p_j$, and $\text{NN}^{-1}(x_i)$ is the set of original points mapped to $x_i$. 

% Thus, regions with a higher density score $\delta$, indicating greater geometric complexity, have smaller drop ratios and retain more channels to preserve information, whereas regions with lower $\delta$, representing simpler geometry, have more channels dropped. 

In SemanticKITTI~\cite{behley2019semantickitti} and ShapeNet~\cite{wu20153d}, point clouds exhibit considerable variability in density and distribution, making fixed normalization parameters inadequate. 
To address this, %we dynamically adapt the normalization parameters $d_{\max}$ and $m_{\max}$ using an Exponential Moving Average (EMA) approach~\cite{ioffe2015batch}:
we adopt the Exponential Moving Average (EMA) approach~\cite{ioffe2015batch}, which dynamically updates the normalization parameters $d_{\max}$ and $m_{\max}$ to reflect the varying density and distance distributions across different datasets. 
Denote the normalization parameters for $d_{\text{num}}$ and $d_{\text{dist}}$ as $\theta \in \{d_{\max}, m_{\max}\}$, and the current training iteration as $t$. 
The normalization parameters $d_{\max}$ and $m_{\max}$ are dynamically updated as: 
\begin{equation}
\theta^{(t)} = (1 - \gamma) \cdot \theta^{(t-1)} + \gamma \cdot P_{95}(m^{(t)}),
\label{eq:ema_update}
\end{equation}
% where $\theta \in \{d_{\max}, m_{\max}\}$ denotes the normalization parameters for $d_{\text{num}}$ and $d_{\text{dist}}$, and $t$ is the current training iteration. 
%As shown in Eq.~\ref{eq:ema_update}, $d_{\max}$ and $m_{\max}$ 
to incorporate both the previous parameters $\theta^{(t-1)}$, and the 95th percentile of current batch statistics $P_{95}(m^{(t)})$\footnote{Experiments across various percentiles (90th-99th) and adaptation rates ($[0.01, 0.5]$) revealed that $P_{95}$ and $\gamma = 0.1$ yield optimal performance.}, enabling gradual adaptation to dataset characteristics while maintaining stability.
%Empirical evaluation across various percentiles (90th-99th) and adaptation rates ($\gamma \in [0.01, 0.5]$) revealed that $P_{95}$ and $\gamma = 0.1$ yield optimal performance. 
% This adaptive strategy ensures consistent performance across diverse point cloud geometries, capturing the upper range of the distribution while mitigating outlier influence.

% \subsubsection{Channel Importance with Gradient Information}
\subsubsection{Global and Local Variance-based Channel Importance}

Point clouds contain complex surface geometries with sharp edges, fine details, and intricate structures, leading to strong spatial variations across feature channels~\cite{qi2017pointnet++, wang2019dynamic}. 
%These variations often encode critical geometric transitions such as edges, corners, and surface detail transitions~\cite{qi2017pointnet++, wang2019dynamic}.
We propose an enhanced channel-importance calculation method that augments variance-based metrics with gradient information % to calculate channel importance 
while maintaining low computational cost. 

Specifically, let the variance feature values in channel $c$ be denoted as $\text{Var}_c$. The gradient metric for the channel, denoted as $\text{Grad}_c$, measures local detail variations 
%by measuring adjacent differences within the channel:
by summarizing the differences between adjacent positions within the channel:
\begin{equation}
\text{Grad}_c = \frac{1}{N-1}\sum_{j=1}^{N-1}|\mathbf{z}_{c,j+1} - \mathbf{z}_{c,j}|,
\label{gradc}
\end{equation}
where $\mathbf{z}_{c,j}$ is the feature value of channel $c$ at position $j$ in the encoded latent space. 

The final channel importance, denoted as $\mathcal{I}_c$, is then computed as a weighted combination of normalized variance and gradient metric in order to capture both globally informative features with high variance and locally discriminative features with high gradient as:
\begin{equation}
\mathcal{I}_c= \beta \cdot \text{norm}(\text{Var}_c) + (1-\beta) \cdot \text{norm}(\text{Grad}_c).
\label{Sc}
\end{equation}
%From Eqs.~\ref{gradc} and~\ref{Sc}, %$N$ denotes the number of points and 
%$\mathbf{z}_{c,j}$ represents the feature value at position $j$ in the encoded latent space. 
%Importance score $\mathcal{I}_c$ combines two normalized statistical measures: $\text{norm}(\text{Var}_c)$ captures global dispersion through standardized sample variance, while $\text{norm}(\text{Grad}_c)$ quantifies local variability. 
$\beta$ is empirically set as $0.6$
%choose the coefficient $\beta = 0.6$ to control 
to balance the relative contribution between global and local components. %, with the normalization function employing max normalization to standardize both measures. 
%This bivariate composite index balances global statistical dispersion with local variability, ensuring that both globally informative features with high variance and locally discriminative features with high gradient can be preserved during coding, making channel selection more robust for diverse geometric structures.

% \subsubsection{Density-aware Combined Dropping Strategy for Features and Coordinates}
\subsubsection{Density-aware Tail Drop}

While ProgDTD~\cite{hojjat2023progdtd} extends the tail-drop~\cite{koike2020stochastic} from a latent feature-only drop strategy to a both latent and hyper-latent~\cite{balle2018variational} drop strategy in 2D image coding, we propose tail-drop for both latent features and down-sampled coordinates to achieve better performance, as confirmed by our Ablation Study in Sect.~\ref{ablation_dropping_strategies}.

Learning-based PCC normally relies on %deals with 
two equally critical data modalities: semantic features that encode %encoding 
surface properties and coordinate features that capture %representing 
spatial geometry~\cite{he2022density}~\cite{guarda2022ipleiria}~\cite{luo2023transformer}. 
% The coordinate information, unlike the hyper-latent in image coding, 
Unlike the hyper-latent in image coding, coordinate information 
constitutes the core geometric data that directly determines %shape 
reconstruction quality. 
% Meanwhile, coordinate features exhibit varying importance across spatial regions, 
Meanwhile, the importance of coordinate features varies across spatial regions, 
with surfaces typically demonstrating high spatial correlation and redundancy. Thus, we propose a density-aware tail drop method that employs a combined drop strategy to achieve this %that addresses this 
by applying 
density-guided progressive coding to %both feature types 
both features and coordinates %simultaneously 
based on their channel importance, to ensure geometric consistency between coordinate and semantic representations. 
Specifically, given a drop ratio $\rho$ and channel importance % rankings 
of features and coordinates, denoted as $\mathcal{I}_z = \{\mathcal{I}_1, \mathcal{I}_2, \ldots, \mathcal{I}_{C_z}\}$ and $\mathcal{I}_{xyz} = \{\mathcal{I}_1^{xyz}, \mathcal{I}_2^{xyz}, \ldots, \mathcal{I}_{C_{xyz}}^{xyz}\}$, respectively, our density-aware tail drop strategy %apply \textcolor{red}{synchronized} channel selection:
retains the top $(1-\rho)$ fraction of channels in both feature and coordinate spaces, as:
\begin{equation}
\mathbf{z}_{\rho} = \mathbf{z} \odot \mathbf{M}(\mathcal{I}_z, \rho), \quad \mathbf{z}_{xyz,\rho} = \mathbf{z}_{xyz} \odot \mathbf{M}(\mathcal{I}_{xyz}, \rho).
\label{eq:both_drop}
\end{equation}
Here, $\mathbf{M}(\mathcal{I}, \rho) \in \{0,1\}^C$ is a binary mask that preserves %retains 
the top $(1-\rho) \times C$ channels according to importance ranking $\mathcal{I}$, $C$ is the total channel count, and $\odot$ denotes element-wise multiplication. 
This synchronized drop with a common %identical 
$\rho$ maintains correspondence between feature and coordinate channels throughout the coding process. 
%This combined dropping strategy offers several distinct advantages over feature-only dropping, including consistent dropping patterns, synchronized gradients, and superior rate-distortion performance. % compared to feature-only dropping strategies.

\subsection{Loss Function}

% For our learning-based point cloud geometry coding framework, we adopt an integrated loss function following the D-PCC~\cite{he2022density} approach. T
Following D-PCC~\cite{he2022density}, we adopt an integrated loss $\mathcal{L}$, which is formulated as a weighted sum of %geometry reconstruction terms and rate constraints:
geometry reconstruction quality terms and coding efficiency constraints:
\begin{equation}\label{loss}
\mathcal{L} = \mathcal{L}_{\text{CD}} + \sigma \cdot \mathcal{L}_{\text{Dens}} + \omega \cdot \mathcal{L}_{\text{Coord}} + \eta \cdot \mathcal{L}_{\text{Points}} + \lambda \cdot \mathcal{R}_{\text{BPP}}.
\end{equation}
Here, $\mathcal{L}_{\text{CD}}$ measures the Chamfer Distance between the original and reconstructed point clouds, ensuring geometric fidelity, $\mathcal{L}_{\text{Dens}}$ preserves local density distributions, $\mathcal{L}_{\text{Coord}}$ regularizes the quantized spatial representation before and after the entropy bottleneck, $\mathcal{L}_{\text{Points}}$ constrains the total number of reconstructed points, and finally, $\mathcal{R}_{\text{BPP}}$ represents the bit rate loss. The weighting coefficients $\sigma$, $\omega$, $\eta$, and $\lambda$ control the trade-off between different reconstruction quality aspects and coding efficiency. Consistent with D-PCC~\cite{he2022density}, we use the same weighting coefficients, enabling fair comparison with baseline methods. 

\begin{figure*}[ht]
    \centering
    \subfloat[SemanticKITTI: BPP vs Chamfer Distance]{\includegraphics[width=0.35\textwidth]{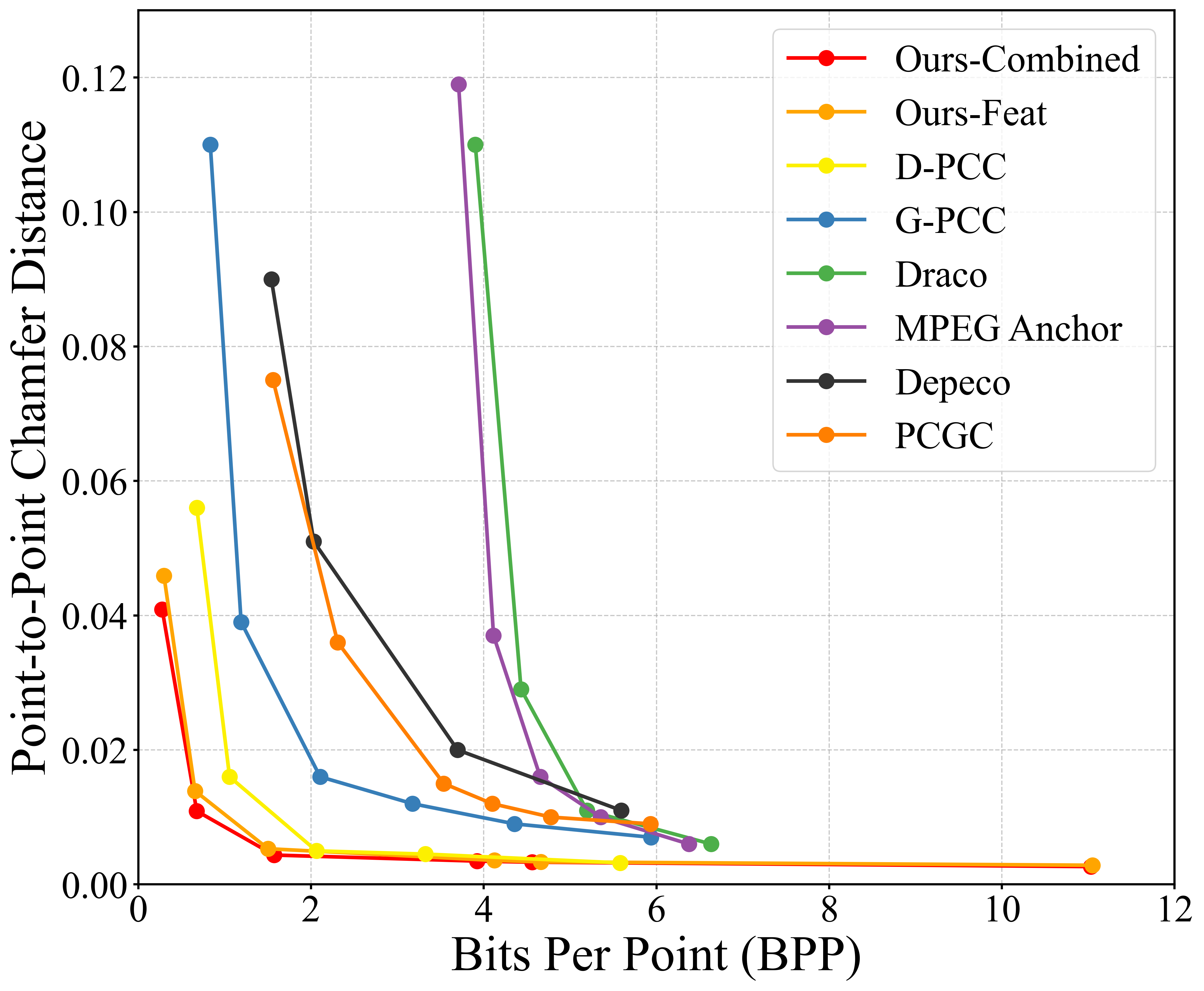}\label{subfig:semantickitti_cd}}
    \hspace{3em} % Add some space between figures
    \subfloat[SemanticKITTI: BPP vs PSNR]{\includegraphics[width=0.35\textwidth]{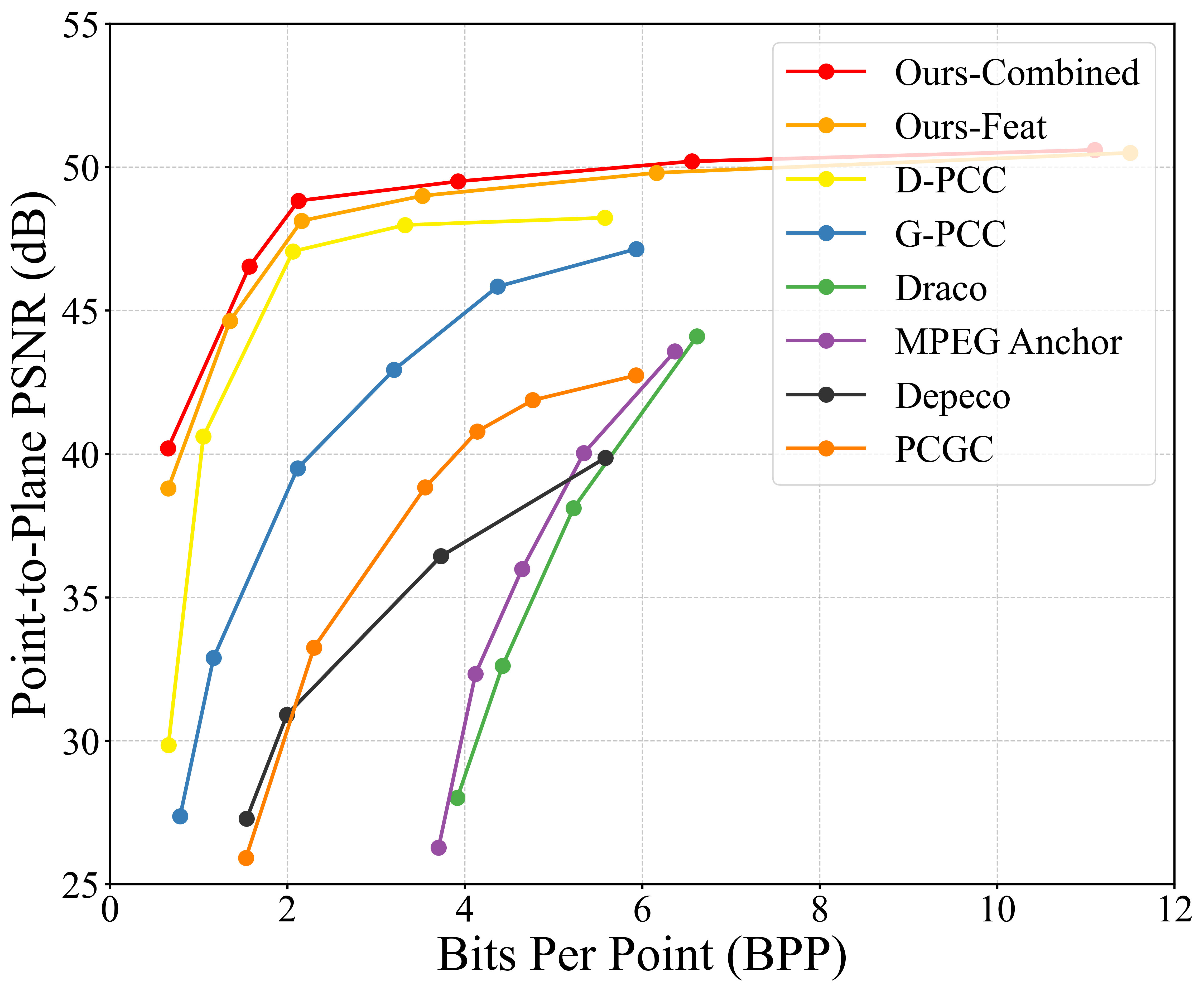}\label{subfig:semantickitti_psnr}}
    \hfill
    \subfloat[ShapeNet: BPP vs Chamfer Distance]{\includegraphics[width=0.35\textwidth]{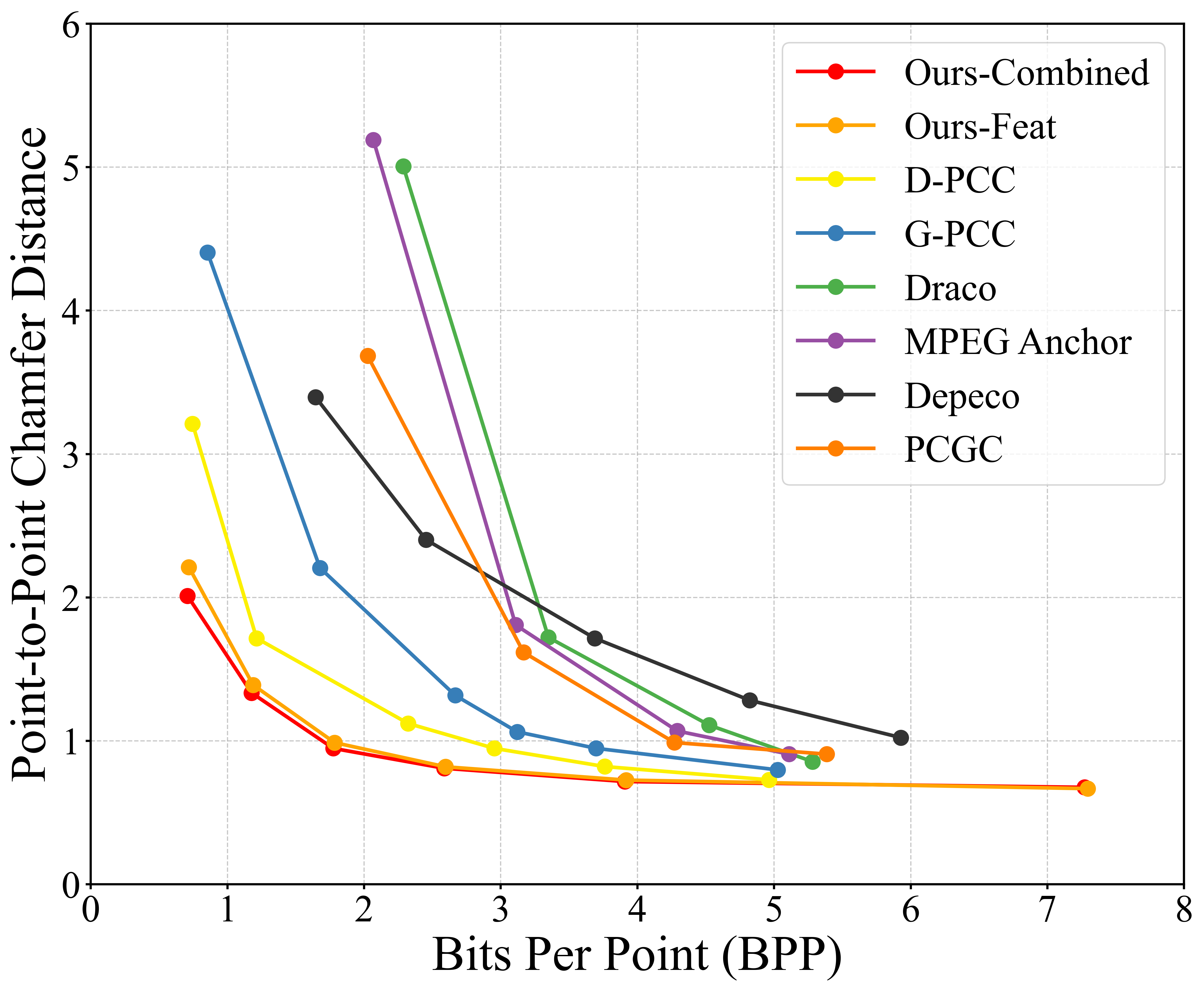}\label{subfig:ShapeNet_cd}}
    \hspace{3em} % Add some space between figures
    \subfloat[ShapeNet: BPP vs PSNR]{\includegraphics[width=0.35\textwidth]{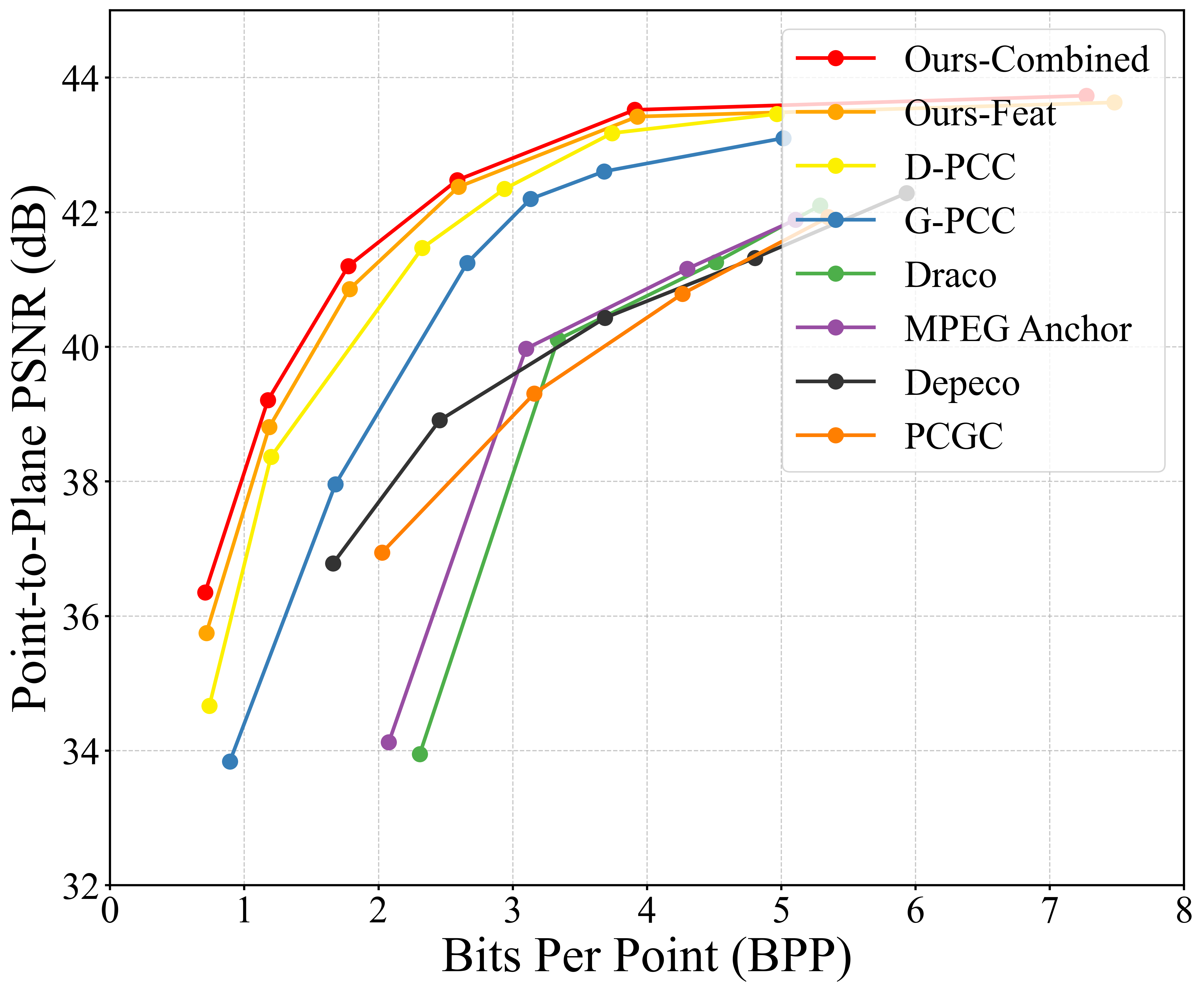}\label{subfig:ShapeNet_psnr}}
    % \caption{Quantitative results of Non-Progressive ProDAT on SemanticKITTI~\cite{behley2019semantickitti} and ShapeNet~\cite{wu20153d}. These figures display BPP versus CD (a and c) and BPP versus PSNR-D2 (b and d) for both datasets. Results compare two strategies: Combined Drop (red lines) and Feature-Only Drop (orange lines), with the model trained using progressive coding.}
    \caption {Quantitative results of non-progressive ProDAT on SemanticKITTI and ShapeNet: BPP vs. CD (a, c) and BPP vs. PSNR-D2 (b, d). %Results compare Combined Drop (red) and Feature-Only Drop (orange), with 
    Models are trained using progressive coding. Results compare two tail-drop strategies: Combined Drop (red lines) and Feature-Only Drop (orange lines), with the model trained using progressive coding. The differences between these strategies will be described in terms of the Ablation Study below.}
    \vspace{-1.5em}
    \label{fig:quantitative_results_1}
\end{figure*}

\subsection{Progressive Coding Evaluation}

\subsubsection{Channel-based Progressive Coding Evaluation}

To evaluate the performance of progressive rate distortion, we conduct %implement a 
channel-based test. % methodology. % consistent with our training framework. 
Given the importance ranking $\mathcal{I}$ computed from latent features and coordinates, we simulate different coding ratios by retaining varying ratios %numbers 
of channels. 
Given a drop ratio $\rho$, the top $k = \lceil(1-\rho) \cdot C\rceil$ channels are reserved, %where $C$ denotes the total channel count. 
and the progressive reconstruction becomes:
\begin{equation}
\hat{\mathbf{z}}_{\rho} = \hat{\mathbf{z}} \odot \mathbf{M}(\mathcal{I_z}, \rho), \quad \hat{\mathbf{z}}_{xyz,\rho} = \hat{\mathbf{z}}_{xyz} \odot \mathbf{M}(\mathcal{I}_{xyz}, \rho),
\end{equation}
where %$\mathbf{M}(\mathcal{I}, \rho)$ generates a binary mask selecting the $k$ highest-importance channels. 
$\hat{\mathbf{z}}$ and $\hat{\mathbf{z}}_{xyz}$ represent quantized latent features and quantized coordinates after entropy bottleneck compression, respectively (see Eq.~\ref{eq:prodat_train}). 
Empirically, we observe a significant %rapid 
quality improvement when $k \in [1, 13]$, with performance plateauing for $k \geq 16$ in our 32-channel configuration.

\subsubsection{Bitrate Calculation}
Following the entropy-based measurement protocols from D-PCC~\cite{he2022density} and H.264 SVC~\cite{wiegand2003overview}, we compute the effective bits-per-point (BPP) for each progressive level. 
Given a drop ratio $\rho$, the BPP is:
\begin{equation}
\mathcal{R}_{\text{BPP}_{\rho}} = \frac{1}{N} \left( \sum_{i \in \mathcal{S}_{\rho}} -\log_2 p(\hat{\mathbf{z}}_i) + \sum_{j \in \mathcal{S}_{xyz,\rho}} -\log_2 p(\hat{\mathbf{z}}_{xyz,j}) \right),
\label{eq:bpp_calc}
\end{equation}
where %$N$ is the total point count, 
$\mathcal{S}_{\rho}$ and $\mathcal{S}_{xyz,\rho}$ denote the sets of retained channel indices for features and coordinates, respectively, and $p(\cdot)$ represents the learned probability distributions from our entropy models $\mathcal{B}_{\mathbf{z}}$ and $\mathcal{B}_{\mathbf{z}_{xyz}}$.

\section{Experiments}
\label {sect:exp}

% In this section, we evaluate the performance of ProDAT. We begin with a brief discussion of implementation details and datasets. Then, we present parameters before focusing on the performance evaluation and conclude this section with a discussion of the results.
% In this section, we evaluate the proposed ProDAT by outlining implementation details and datasets, presenting parameters, analyzing performance, and concluding with a discussion.

\subsection{Datasets and Implementation}

%To evaluate our ProDAT method, we utilize 
%We adopt two benchmark datasets: 
To evaluate the performance of the proposed ProDAT, we conducted experiments and ablation studies on two benchmark datasets, \textit{i.e.}, SemanticKITTI~\cite{behley2019semantickitti} and ShapeNet~\cite{wu20153d}. %ShapeNetCore.v1~\cite{wu20153d}. 
SemanticKITTI consists of approximately 43,000 LiDAR scan frames from urban and suburban driving environments, each containing around 120,000 points with higher density near vehicles due to sensor proximity. ShapeNet~\cite{wu20153d} comprises 51,300 synthetic 3D models across 55 categories, with each model averaging 55,000 points. 
% Our model architecture includes an encoder with three downsampling stages, configurable with factors of 1/2, 1/3, or 1/4, and a decoder with a maximum upsampling factor of 8. 
When preprocessing these datasets for training and testing, we strictly adhere to the requirements specified in D-PCC~\cite{he2022density}. %exactly follow D-PCC~\cite{he2022density} requirements. 
The training objective combines distortion loss, density loss (initialized at \(10^{-4}\)), and coordinate loss (initialized at \(5 \times 10^{-5}\)). Optimization is performed over 50 epochs using the Adam optimizer, with an initial learning rate of \(10^{-3}\), decayed by 0.5 every 15 epochs. All experiments are conducted with an NVIDIA A40 GPU. % for efficient processing of large-scale point clouds.

%Based on experimental results, we increased the number of latent feature channels from the original 8~\cite{he2022density} to 32 to achieve effective progressive coding. This adjustment provides sufficient channel diversity to enhance representation capacity and coding efficiency. In contrast, configurations with 16 channels exhibited inadequate diversity, while 64 channels introduced redundancy without commensurate performance gains, making 32 channels the optimal choice for balancing quality and efficiency. We believe that this improvement in channel capacity not only enables progressive coding but also improves reconstruction quality, achieving lower Chamfer distance loss at equivalent bitrates. Thus, we consistently use this 32-channel configuration, as it provides an optimal balance between coding efficiency and reconstruction quality throughout all experiments. Furthermore, while the original D-PCC~\cite{he2022density}  achieves the best performance with batch size 1 in their model training, we train ProDAT with a batch size of 32 to manage the high computational requirements and time. In other words, we suspect that our model could perform better if had available processing power to  use a batch size of 1. This represents a practical trade-off that our model shows stronger performance despite this batch size, and theoretically, even better results could be achieved with a batch size of 1 if computational resources permit.

Based on experimental results, we increased the number of latent feature channels from the original 8 in~\cite{he2022density} to 32 to enable effective progressive coding. This provides sufficient channel diversity, enhancing representation capacity and coding efficiency: 16 channels yielded inadequate diversity, while 64 introduced redundancy without commensurate gains, making 32 optimal for quality-efficiency balance. 
%This improvement not only supports progressive coding but also enhances reconstruction quality, yielding lower Chamfer distance loss at equivalent bitrates. 
We thus adopted 32 channels across all experiments. 
Further, unlike D-PCC~\cite{he2022density}, which optimizes at batch size 1, we used 32 for ProDAT to reduce computational demands. %—a practical trade-off, as batch size 1 could yield even better results if resources permit, given our model's already superior performance.

\subsection{Evaluation Metrics}

As there is currently no standardization or benchmark for progressive point cloud geometry coding, we compare our results against SOTA learning-based point cloud coding models, %which utilize the same datasets evaluated 
on the same datasets, including D-PCC~\cite{he2022density}, G-PCC~\cite{wallace1992jpeg}, Google Draco~\cite{galligan2018google}, MPEG Anchor~\cite{mekuria2016design}, PCGC~\cite{wang2021lossy}, Depeco~\cite{wiesmann2021deep}, and JPEG Pleno~\cite{guarda2022ipleiria}.

%To comprehensively evaluate the performance of ProDAT, 
We evaluate ProDAT using a suite of metrics: PSNR-D1, PSNR-D2, Chamfer Distance (CD), and Bjontegaard Rate (BD-Rate)~\cite{perry2021jpeg,perry2020jpeg}, to collectively assess perceptual quality, geometric fidelity, and coding efficiency. 

\begin{figure}[!ht]
    \centering
    \subfloat[BPP vs PSNR-D1]{\includegraphics[width=0.23\textwidth]{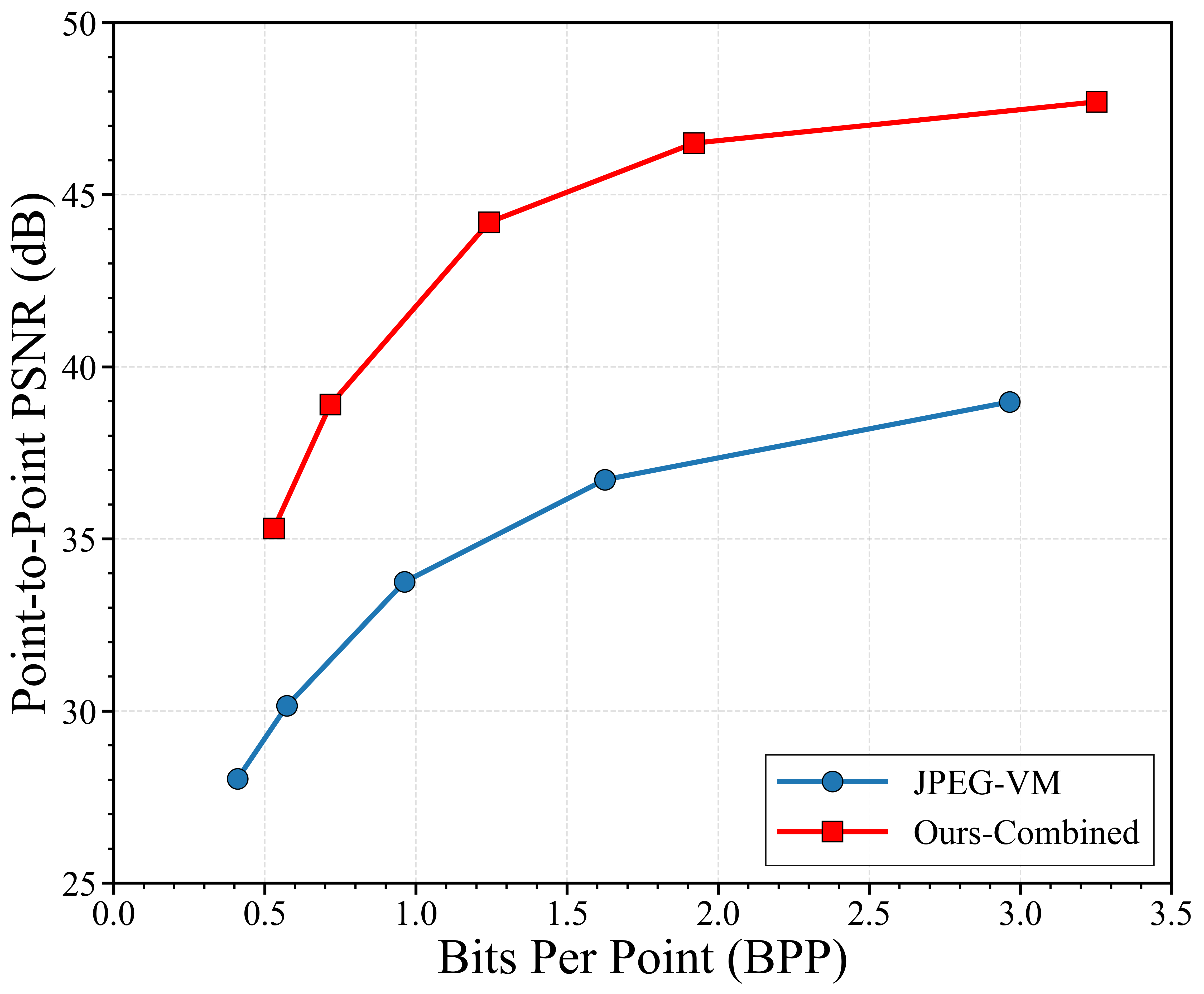}\label{subfig:semantickitti_vm_compare_1}}
    \quad % Add some space between figures
    \subfloat[BPP vs PSNR-D2]{\includegraphics[width=0.23\textwidth]{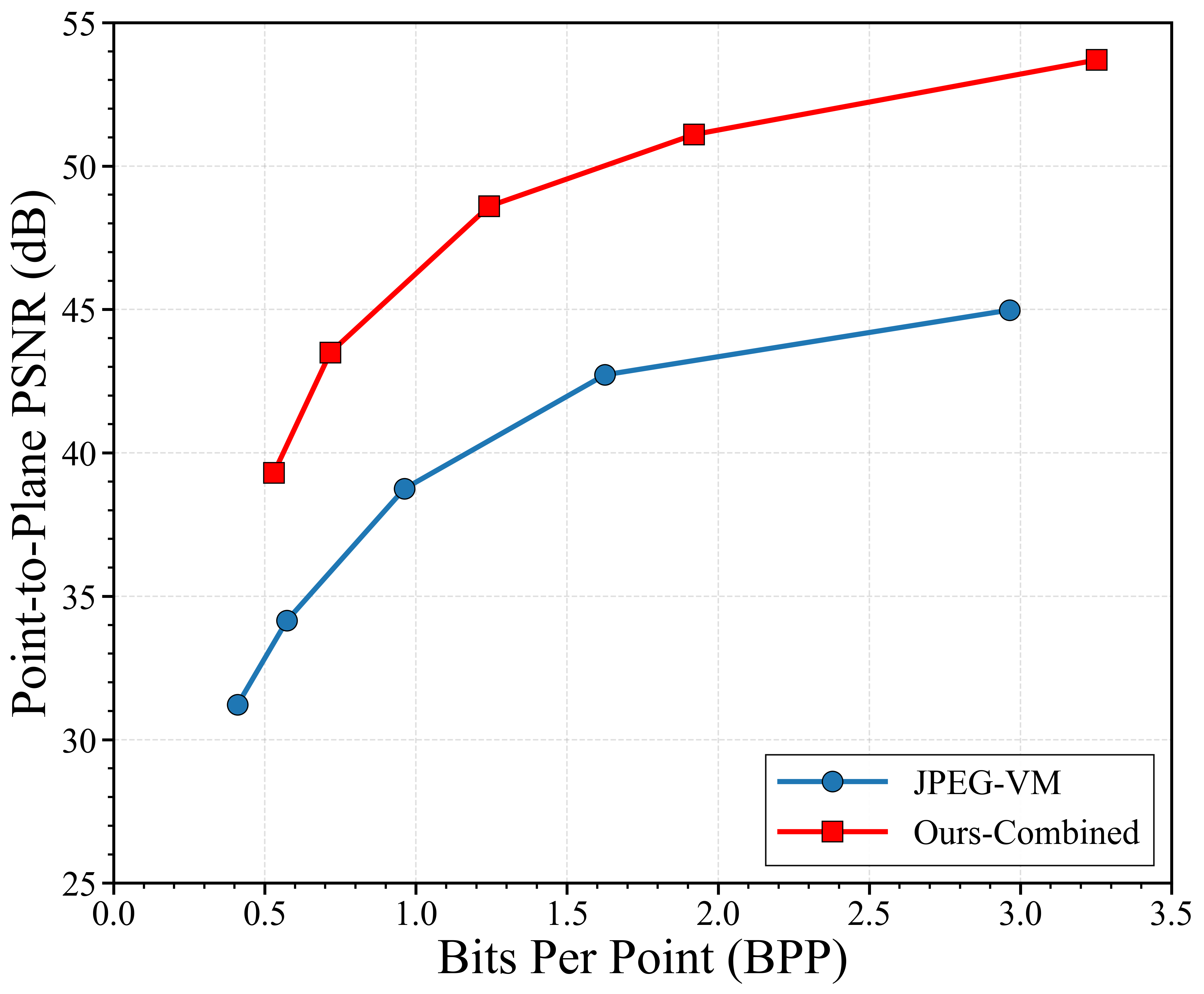}\label{subfig:semantickitti_vm_compare_2}}
    % \caption{Quantitative results of non-progressive ProDAT compared with JPEG-VM~\cite{guarda2022ipleiria} on SemanticKITTI~\cite{behley2019semantickitti}, including PNSR-D1 and PSNR-D2 for each point cloud model in the dataset.}
    \caption{Quantitative comparison of non-progressive ProDAT with JPEG-VM~\cite{guarda2022ipleiria} on SemanticKITTI in terms of PSNR-D1 and PSNR-D2 for each point cloud model.}
    \label{fig:quantitative_results_vm}
    \vspace{-1em}
\end{figure}

\textbf{PSNR-D1} and \textbf{PSNR-D2} are variants of the peak signal-to-noise ratio (PSNR) %tailored for point-cloud quality assessment:  evaluation, 
measuring the quality of a reconstructed point cloud relative to the original. Specifically,
\begin{equation}
\text{PSNR-D} = 10 \cdot \log_{10}\left(\frac{3 \cdot \text{Peak}_{D}^2}{\text{MSE}_{\text{max}}^{(D)}}\right),
\end{equation}
where \( D \) denotes the distance metric. % (D1 or D2). 
The peak signal value, \(\text{Peak}_{D}\), is defined 
% as the squared diagonal of the point cloud's bounding box, 
as the squared length of the bounding-box diagonal, \textit{i.e.}, \(\|\mathbf{p}_{\max} - \mathbf{p}_{\min}\|_2^2\), with \(\mathbf{p}_{\max}\) and \(\mathbf{p}_{\min}\) being the maximum and minimum coordinates of the original point cloud, as in D-PCC~\cite{he2022density}. 
%This value is consistent across both PSNR-D1 and PSNR-D2 for uniformity in evaluation. 
The same \(\mathrm{Peak}_D\) is used for both PSNR-D1 and PSNR-D2 to ensure comparability. 
${\text{MSE}_{\text{max}}^{(D)}}$ is the maximum of two directional mean squared errors (MSE) between the original and reconstructed point clouds:
\begin{equation}
\text{MSE}_{\text{max}}^{(D)} = \max\left( \text{MSE}_{o \to r}^{(D)}, \text{MSE}_{r \to o}^{(D)} \right),
\end{equation}
where \(\text{MSE}_{o \to r}^{(D)}\) is %the mean-squared error from the original to the reconstructed point cloud, 
computed from the original to the reconstructed, 
and \(\text{MSE}_{r \to o}^{(D)}\) vice versa, %is the reverse, 
both calculated with distance metric \( D \). 
For PSNR-D1, \( D \) uses Euclidean nearest-neighbor distances to assess geometric accuracy; For PSNR-D2, \( D \) uses distances projected along surface normals to capture perceptual distortions and surface smoothness, %which is 
critical for applications such as virtual reality.

\textbf{CD} evaluates geometric fidelity and density preservation via the average bidirectional nearest-neighbor distances, 
% serving as a robust tool for shape comparison and as a loss function in 3D tasks~\cite{lin2024infocd, fan2017point}.
and is widely used for shape comparison and as a loss in 3D learning tasks~\cite{lin2024infocd,fan2017point}. 
\textbf{BD-Rate} quantifies coding efficiency 
% by measuring bitrate differences between two coding algorithms at equivalent quality levels, typically based on PSNR or similar quality indicators. A lower BD-Rate indicates better coding performance. 
as the relative bitrate difference between two rates at matched quality (typically using PSNR or a similar metric); lower values indicate better performance. 
%Together, these metrics form a comprehensive evaluation framework, supporting the analysis and development of point cloud processing technologies across diverse 3D applications.

\subsection{Evaluation Results}
\subsubsection{Rate Distortion Evaluation}

Results demonstrate the superior RD performance on the benchmark datasets SemanticKITTI~\cite{behley2019semantickitti} and ShapeNet~\cite{wu20153d}. As shown in Fig.~\ref{fig:quantitative_results_1}, our approach consistently outperforms SOTA methods in quantitative evaluations using CD, PSNR-D1, and PSNR-D2 metrics.

To comprehensively evaluate our method, we report non-progressive results from training and testing without channel drop. 
Notably, ProDAT consistently achieves significantly lower CD values across all bitrates, especially in the low-bitrate range of 0.5--2.0 BPP. For instance, on SemanticKITTI (Fig.~\ref{subfig:semantickitti_psnr}), it attains $\sim$ 43 dB PSNR-D2 at 1.0 BPP, a quality level that competing methods only reach at substantially higher bitrates. Similarly, on ShapeNet (Fig.~\ref{subfig:ShapeNet_psnr}), it achieves $\sim$ 39.2 dB at 2.0 BPP, underscoring its efficiency in preserving geometric fidelity. 
%Moreover, our method extends the bitrate range beyond D-PCC~\cite{he2022density}: 
Besides quality improvements, ProDAT also extends the supported larger bitrate range than D-PCC~\cite{he2022density}: 
from 0--6 BPP to 0--11 BPP on SemanticKITTI and 0--5 BPP to 0--7 BPP on ShapeNet, enhancing flexibility for bandwidth-limited applications. 
Moreover, the BD-rate metric shows that our combined Drop method achieves 28.6\% bitrate savings on SemanticKITTI and 18.15\% on ShapeNet relative to D-PCC~\cite{he2022density}, at equivalent quality levels. 
This consistent performance across structured outdoor environments and object-centric data highlights the versatility and robustness of our method for progressive point cloud coding.

\begin{figure}[!t]
    \centering
    \subfloat[SemanticKITTI]%: BPP vs PSNR-D2 Progressive Compression with various lambda and both drops]
    {\includegraphics[width=0.23\textwidth]{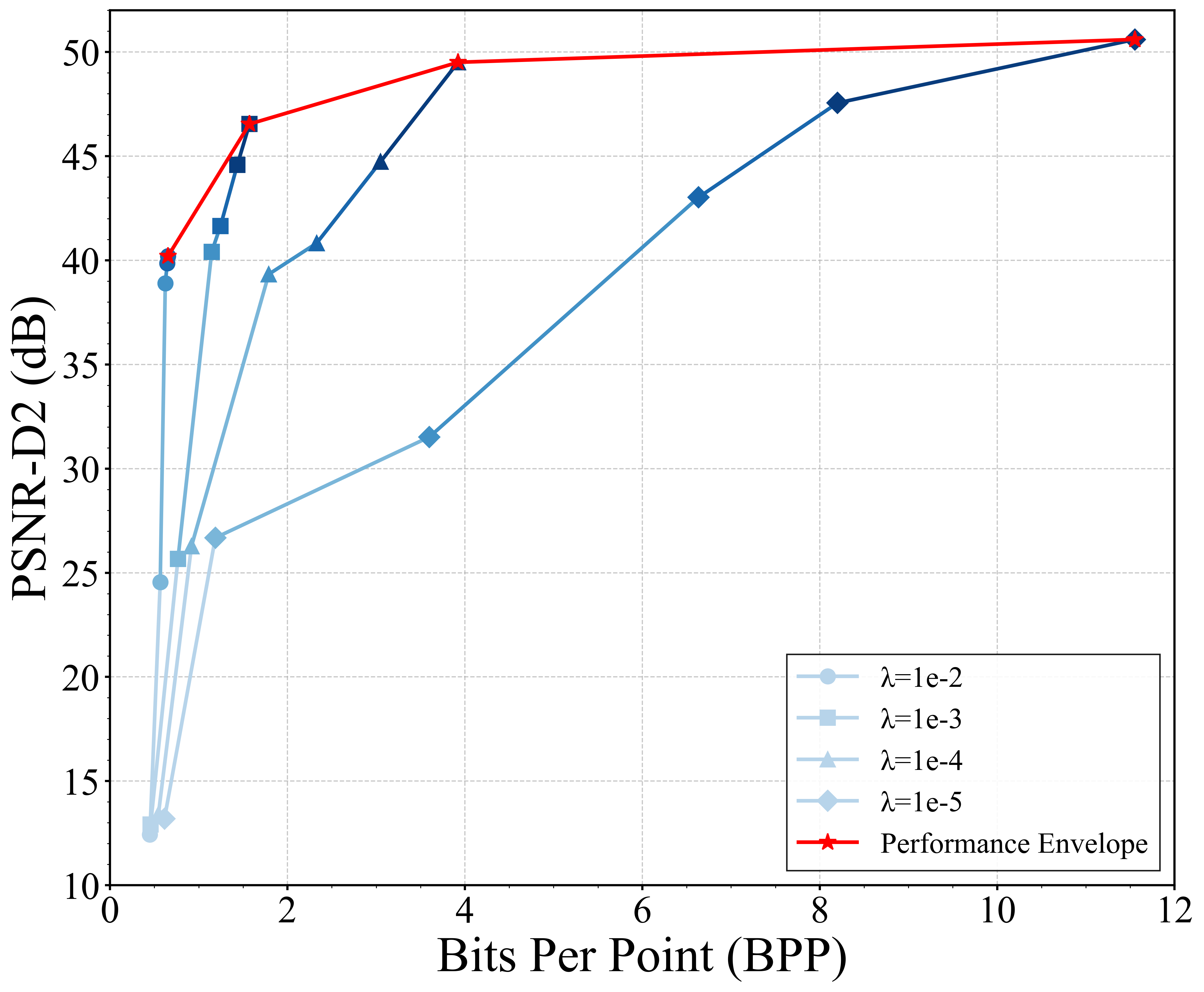}\label{subfig:semantickitti_progressive_both}}
    \quad % Add some space between figures
    \subfloat[ShapeNet]%: BPP vs PSNR-D2 Progressive Compression with various lambda and both drops]
    {\includegraphics[width=0.23\textwidth]{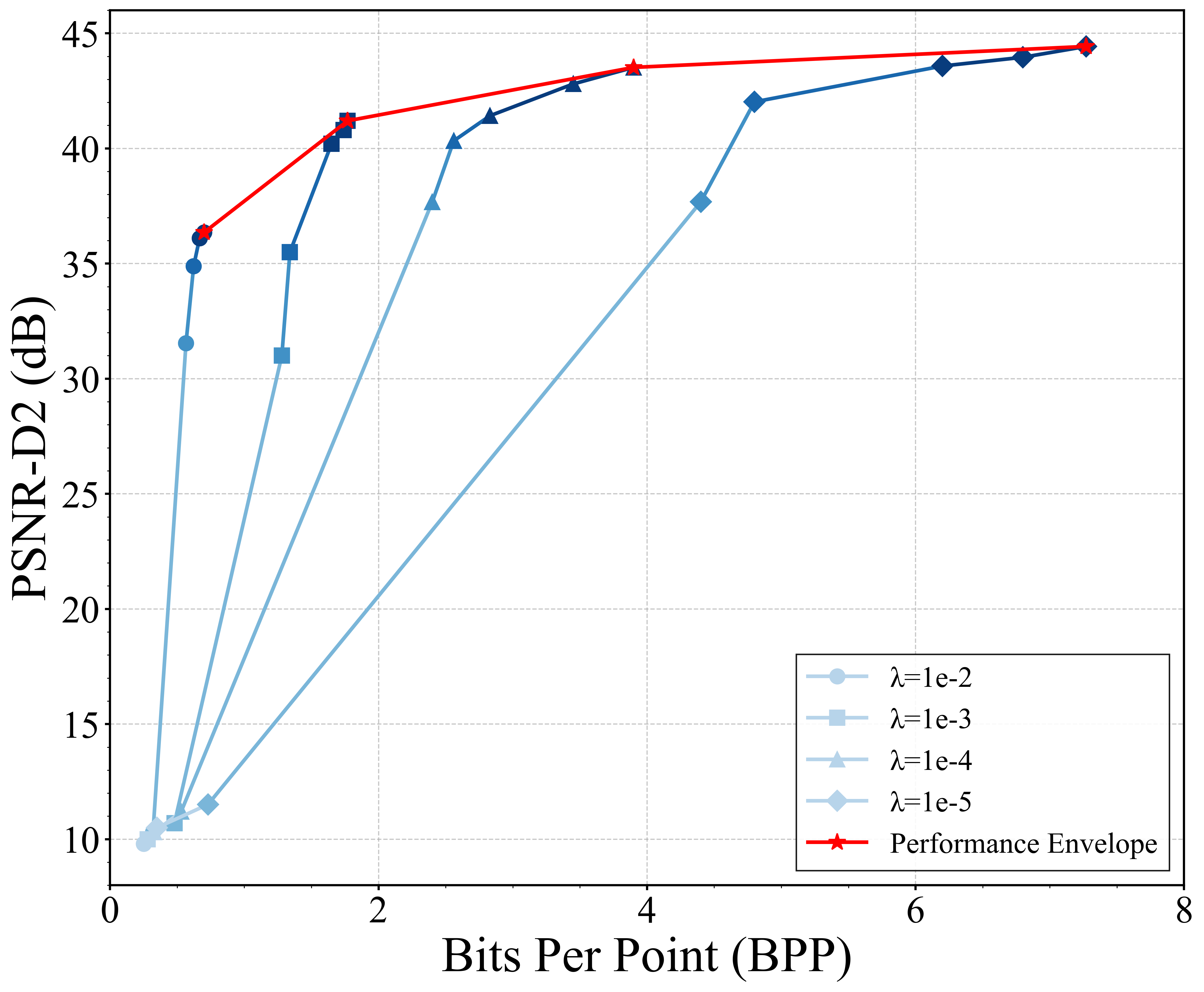}\label{subfig:ShapeNet_progressive_both}}
    % \caption{Progressive performance with both latent feature and coordinate drop on PSNR-D2 from ProDAT in SemanticKITTI and ShapeNet with various lambda.}
    % \caption{Progressive performance of ProDAT (PSNR-D2 vs BPP) with simultaneous latent feature and coordinate drop, evaluated on SemanticKITTI and ShapeNet for different $\lambda$ values.}
    \caption{Progressive performance of ProDAT (PSNR-D2 vs BPP) on SemanticKITTI and ShapeNet across different $\lambda$ values with simultaneous latent feature and coordinate drop.}
    \vspace{-1.5em}
    \label{fig:quantitative_results}
\end{figure}

To benchmark our ProDAT model against the JPEG-VM standard~\cite{guarda2022ipleiria} for learning-based PCC, we utilized the SemanticKITTI dataset~\cite{behley2019semantickitti}. 
%The dataset was voxelized 
Point clouds were voxelized at a 10-bit precision %(resolution of 1023) 
to balance geometric fidelity and coding efficiency. 
Each LiDAR point cloud was normalized to a unit cube via scene-adaptive bounding boxes, with outliers removed and 5\% padding to reduce boundary artifacts, and then quantized to integers in [0, 1023]. 
The preprocessed dataset was evaluated using checkpoints from both ProDAT and JPEG-VM~\cite{guarda2022ipleiria}. 
As shown in Fig.~\ref{fig:quantitative_results_vm}, ProDAT achieves substantial gains over JPEG-VM, with BD-rate reductions exceeding 65\% for PSNR-D1 and 55\% for PSNR-D2 on SemanticKITTI. 
% The primary factor contributing to the observed BD-rate savings is the utilization of pre-trained models provided for the JPEG-VM model~\cite{guarda2022ipleiria} during evaluation on the SemanticKITTI dataset~\cite{behley2019semantickitti} without retraining.
These BD-rate savings are largely attributable to evaluating JPEG-VM with publicly released pre-trained models on SemanticKITTI without retraining.

\subsubsection{Progressive Performance on PSNR-D2}% from ProDAT}

\begin{figure*}[ht]
    \centering

    \makebox[0.18\textwidth][c]{\small Ground Truth}
    \hspace{0.05in}
    \makebox[0.18\textwidth][c]{\small PR=0.03}
    \hspace{0.05in}
    \makebox[0.18\textwidth][c]{\small PR=0.09}
    \hspace{0.05in}
    \makebox[0.18\textwidth][c]{\small PR=0.3}
    \hspace{0.05in}
    \makebox[0.18\textwidth][c]{\small PR=0.5}
    % \vspace{0.2cm}

    \includegraphics[width=0.18\textwidth]{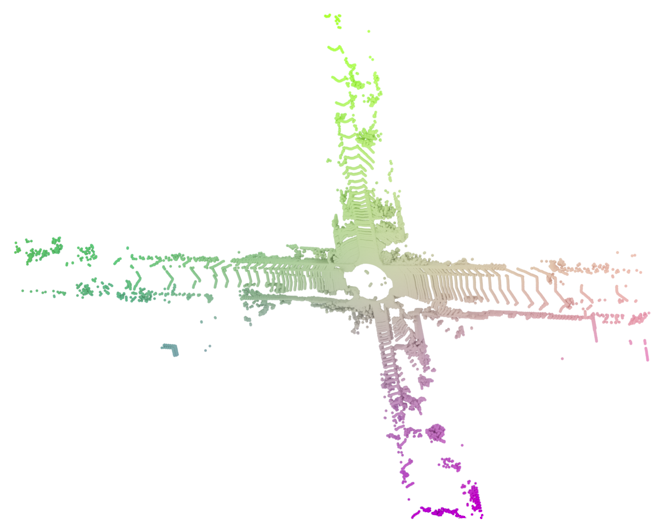}
    \hspace{0.05in}
    \includegraphics[width=0.18\textwidth]{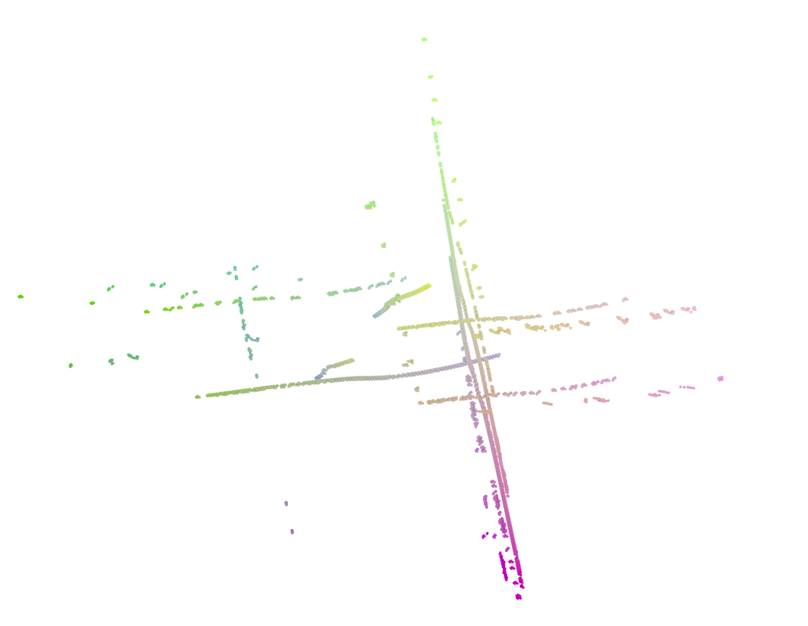}
    \hspace{0.05in}
    \includegraphics[width=0.18\textwidth]{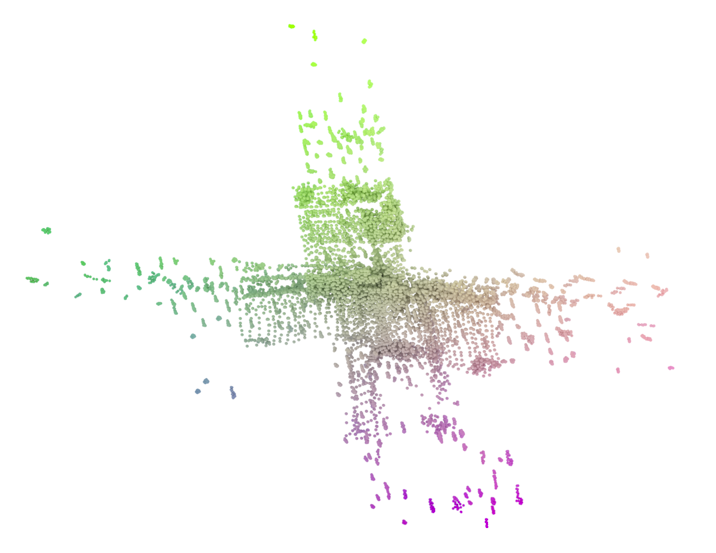}
    \hspace{0.05in}
    \includegraphics[width=0.18\textwidth]{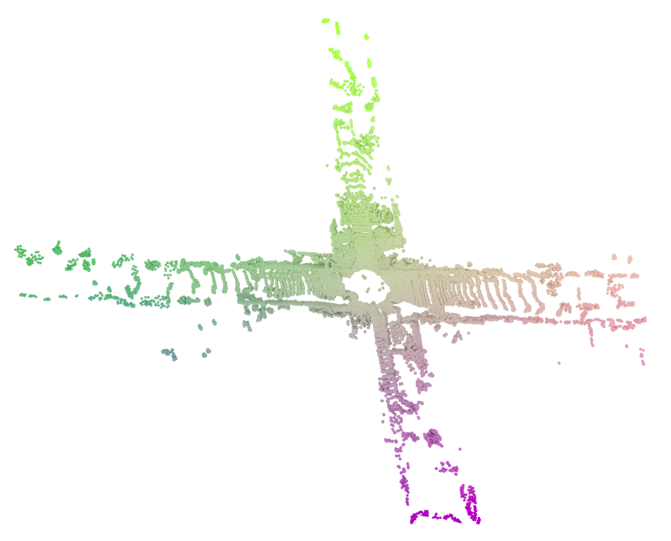}
    \hspace{0.05in}
    \includegraphics[width=0.18\textwidth]{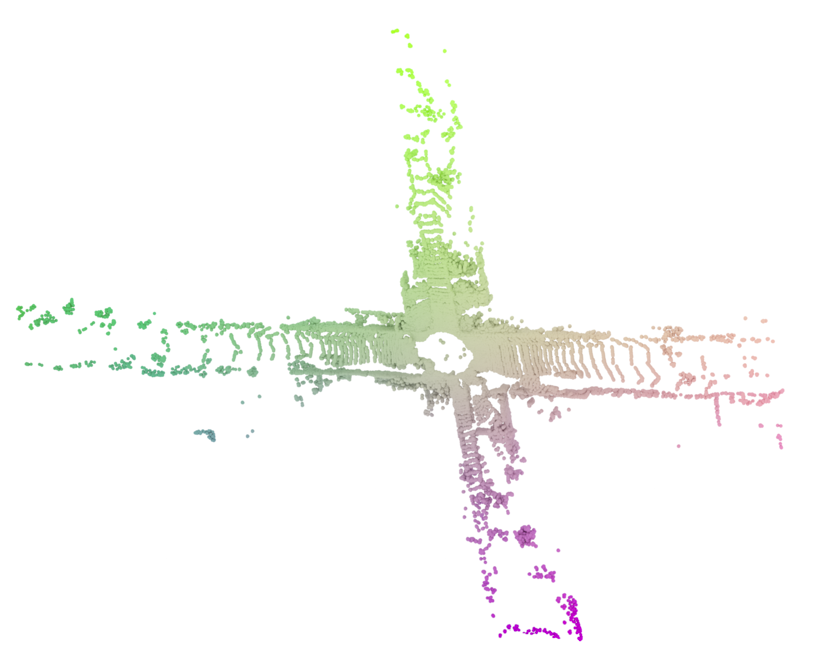}
    % \vspace{0.2cm} % Vertical space after the row

    \includegraphics[width=0.18\textwidth]{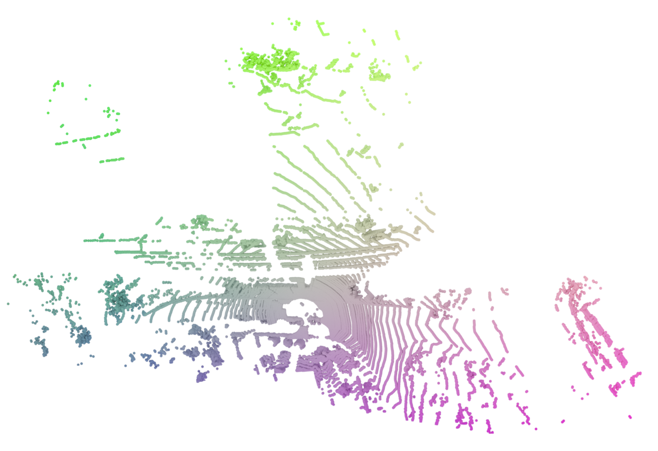}
    \hspace{0.05in}
    \includegraphics[width=0.18\textwidth]{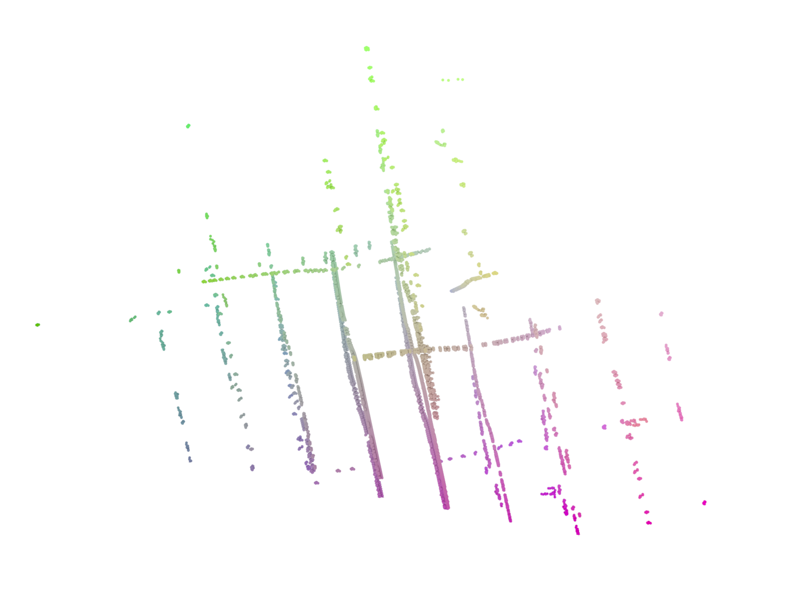}
    \hspace{0.05in}
    \includegraphics[width=0.18\textwidth]{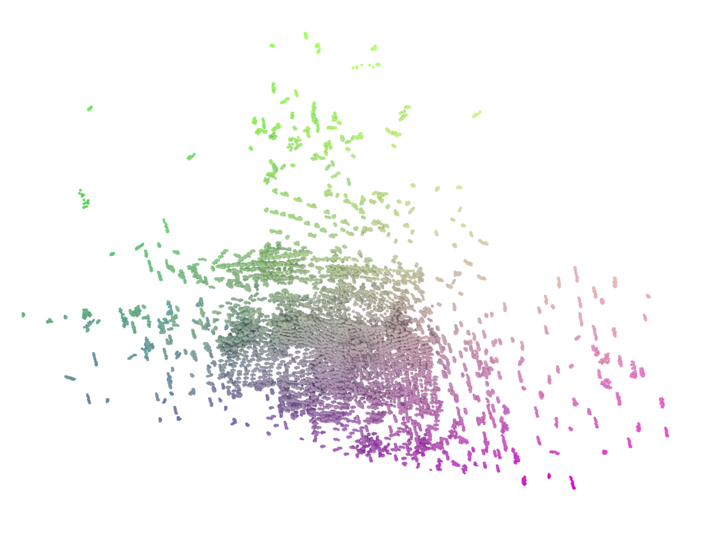}
    \hspace{0.05in}
    \includegraphics[width=0.18\textwidth]{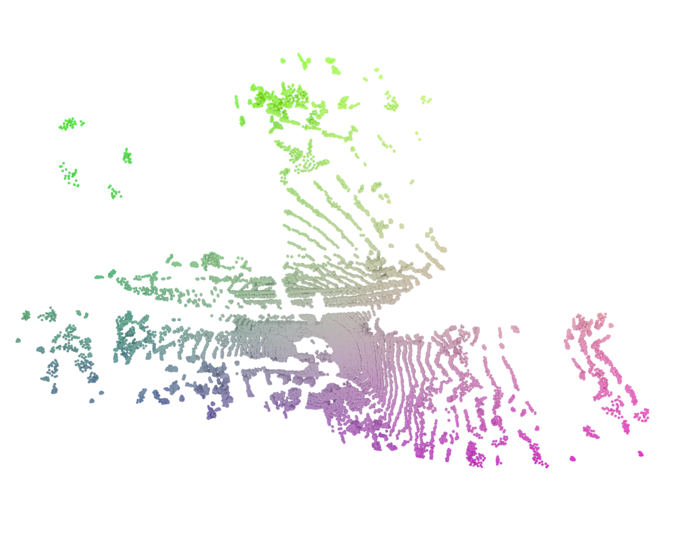}
    \hspace{0.05in}
    \includegraphics[width=0.18\textwidth]{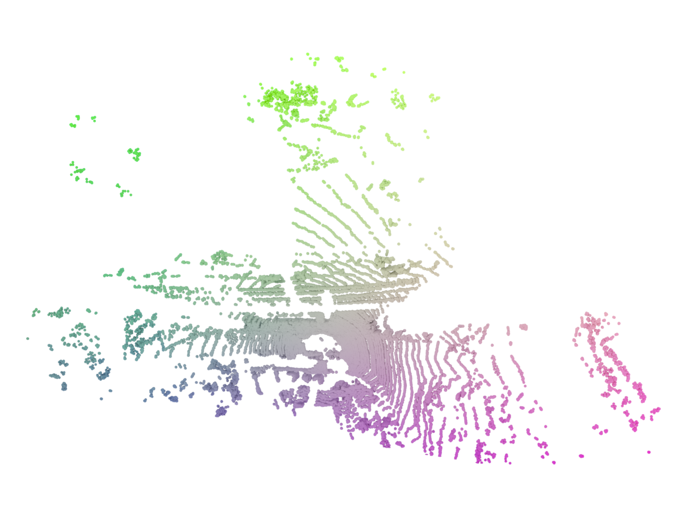}

    % \vspace{0.5cm} % Original vertical space before the caption

    % \caption{Visualization of the progressive coding applied to two SemanticKITTI models~\cite{behley2019semantickitti}, illustrating the impact of Combined-Drop and Feature-only-Drop. From left to right: Ground Truth, PR=0.03, 0.09, 0.3, and 0.5, corresponding to drop rates (1-PR) of 0.97, 0.91, 0.7, and 0.5, respectively.}
    \caption{Visualization of the progressive coding applied to two SemanticKITTI models~\cite{behley2019semantickitti} with various PR values, illustrating the impact of Combined-Drop. }
    \label{fig:semantickitti_compression_viz} % Added a label for easier referencing
    \label{fig7}
    \vspace{-1em}
\end{figure*}

\begin{figure*}[ht]
    \centering

    \makebox[0.18\textwidth][c]{\small Ground Truth}
    \hspace{0.05in}
    \makebox[0.18\textwidth][c]{\small PR=0.09}
    \hspace{0.05in}
    \makebox[0.18\textwidth][c]{\small PR=0.12}
    \hspace{0.05in}
    \makebox[0.18\textwidth][c]{\small PR=0.19}
    \hspace{0.05in}
    \makebox[0.18\textwidth][c]{\small PR=0.25}
    % \vspace{0.2cm}

    \includegraphics[width=0.18\textwidth]{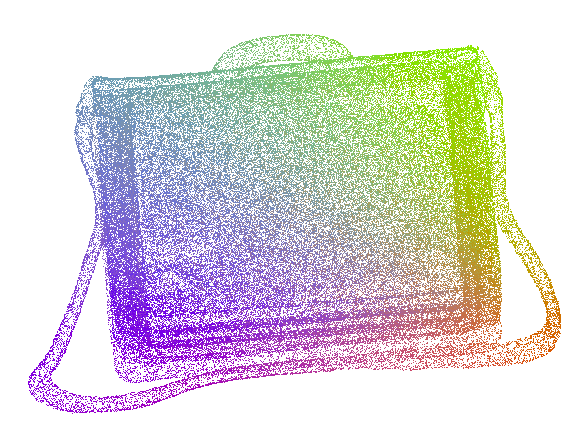}
    \hspace{0.05in}
    \includegraphics[width=0.18\textwidth]{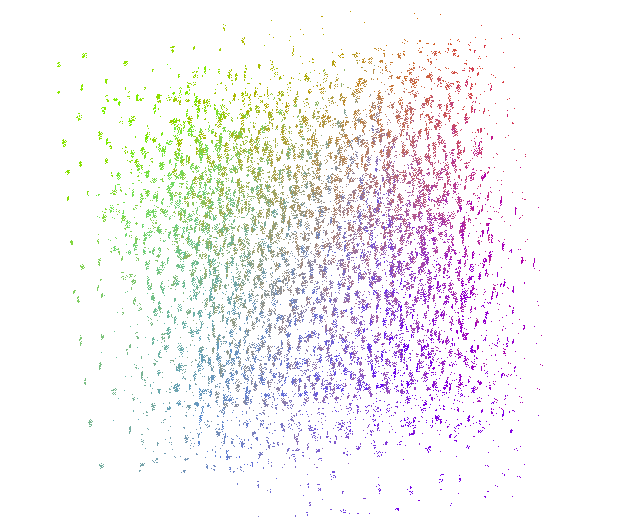}
    \hspace{0.05in}
    \includegraphics[width=0.18\textwidth]{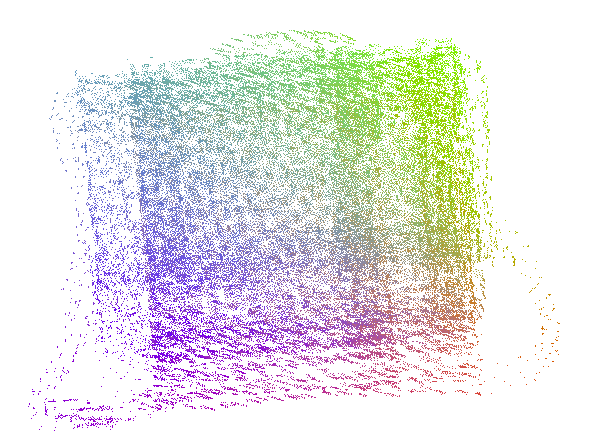}
    \hspace{0.05in}
    \includegraphics[width=0.18\textwidth]{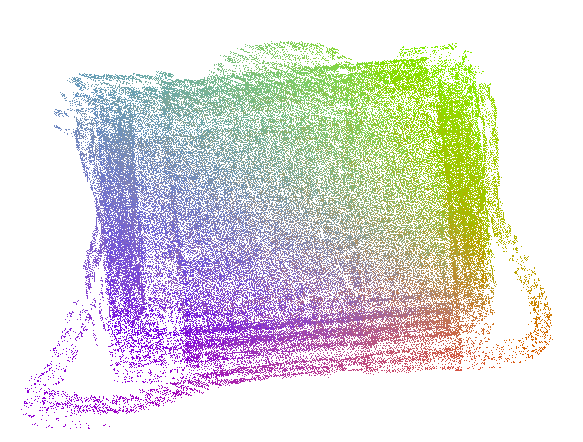}
    \hspace{0.05in}
    \includegraphics[width=0.18\textwidth]{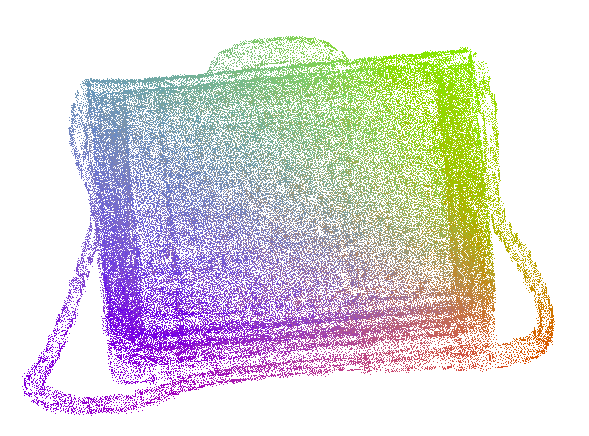}
    % \vspace{0.2cm} % Vertical space after the row

    % \includegraphics[width=0.18\textwidth]{figures/vis/shapenet/model34_org.png}
    % \hspace{0.05in}
    % \includegraphics[width=0.18\textwidth]{figures/vis/shapenet/model34_ratio_009.png}
    % \hspace{0.05in}
    % \includegraphics[width=0.18\textwidth]{figures/vis/shapenet/model34_ratio_012.png}
    % \hspace{0.05in}
    % \includegraphics[width=0.18\textwidth]{figures/vis/shapenet/model34_ratio_019.png}
    % \hspace{0.05in}
    % \includegraphics[width=0.18\textwidth]{figures/vis/shapenet/model34_ratio_025.png}
    % \vspace{0.2cm} % Vertical space after the row

    \includegraphics[width=0.18\textwidth]{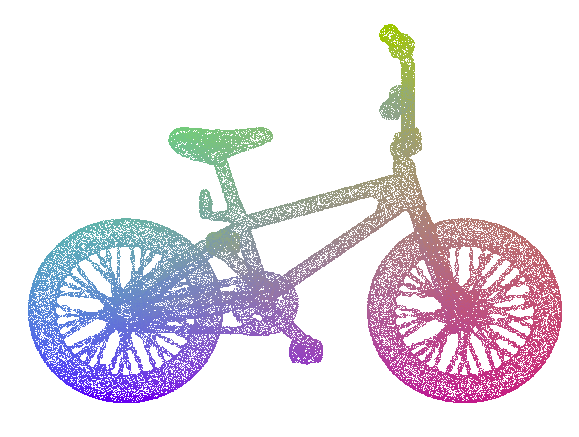}
    \hspace{0.05in}
    \includegraphics[width=0.18\textwidth]{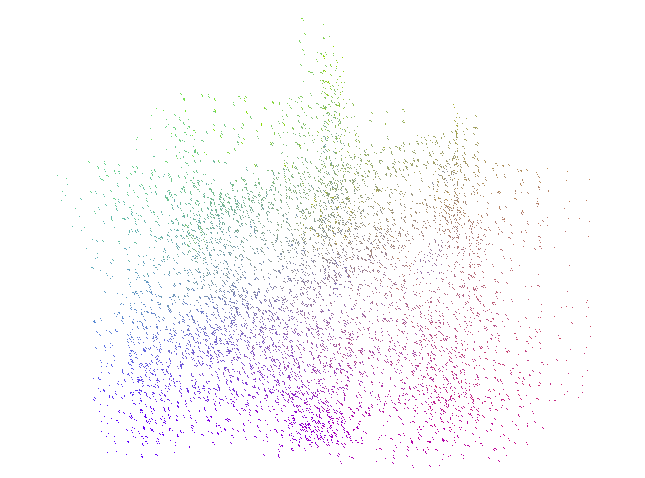}
    \hspace{0.05in}
    \includegraphics[width=0.18\textwidth]{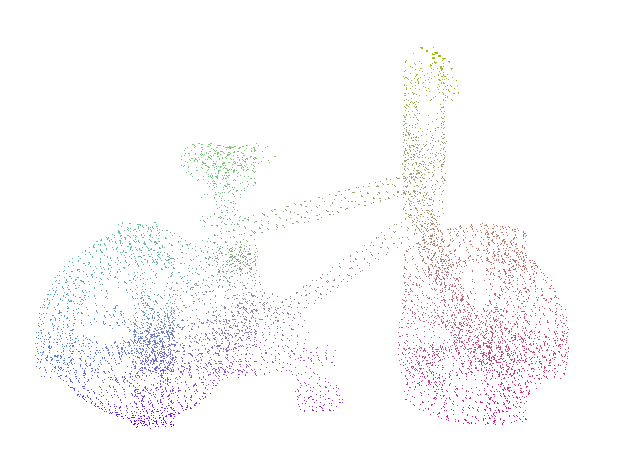}
    \hspace{0.05in}
    \includegraphics[width=0.18\textwidth]{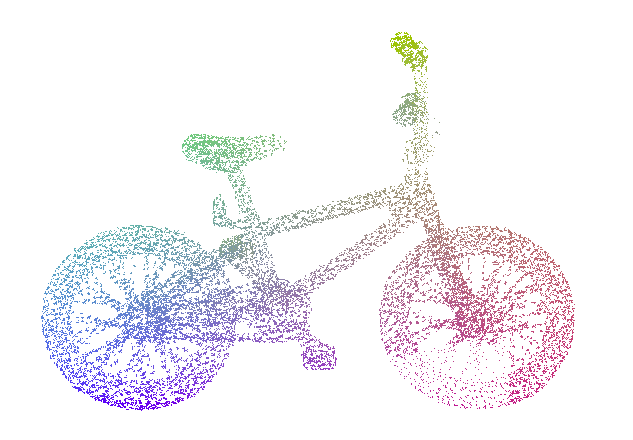}
    \hspace{0.05in}
    \includegraphics[width=0.18\textwidth]{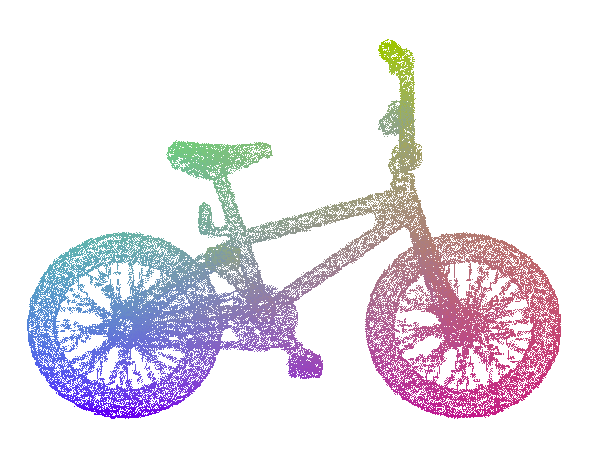}
    % \vspace{0.2cm} % Vertical space after the row

    % \vspace{0.5cm} % Original vertical space before the caption

    \caption{Visualization of the progressive coding applied to two ShapeNet models~\cite{wu20153d} with lambda as 0.001, illustrating the impact of combined feature and coordinate drop. %To achieve more precise visualizations, we set the overlap ratio to 0.2 during dataset preprocessing, such as the ground truth, while keeping all other preprocessing steps unchanged, and we use the model trained from a non-overlapping dataset to test it; the result is still solid. 
    From left to right: Ground Truth, Progressive Ratio (PR) = 0.09, 0.12, 0.19, and 0.25, corresponding to drop rates (1-PR) of 0.91, 0.88, 0.81, and 0.75, respectively. }
    \label{fig:ShapeNet_compression_viz} % Added a label for easier referencing
    \vspace{-1em}
    \label{fig8}
\end{figure*}

% We focus on the PSNR-D2 metric to assess the progressive coding performance of our model, following the de facto benchmark for point cloud compression evaluation in recent literature~\cite{he2022density}. 
We evaluate progressive coding using the \textbf{PSNR-D2} metric, following the de facto benchmark for point-cloud compression~\cite{he2022density}. 
Figs.~\ref{subfig:semantickitti_progressive_both} and~\ref{subfig:ShapeNet_progressive_both} show the rate--distortion (RD)  trade-off for SemanticKITTI and ShapeNet, respectively. 
Each plot includes multiple curves for different values of the trade-off parameter \(\lambda\) that controls the balance between coding efficiency and reconstruction quality. 
%Additionally, a performance envelope is shown, 

In detail, on SemanticKITTI (Fig.~\ref{subfig:semantickitti_progressive_both}), 
%PSNR-D2 exhibits a relatively high initial value at low BPP, ranging from 0 to 1, followed by a gradual increase. 
PSNR-D2 starts relatively high even at low bits (0--1 BPP) and increases gradually; 
Once BPP exceeds 1, %PSNR-D2 demonstrates a more rapid improvement, approaching its maximum value for BPP greater than 4--6. 
it rises more steeply and tends to saturate around 4--6 BPP.
PSNR-D2 starts relatively high even at low bit budgets (0--1 BPP) and increases gradually; once BPP exceeds \(\sim\!1\), it rises more steeply and tends to saturate around 4--6 BPP.
Conversely, on ShapeNet (Fig.~\ref{subfig:ShapeNet_progressive_both}),
%starts with a lower PSNR-D2 at low BPP and shows a slower initial increase. 
PSNR-D2 begins lower at small BPP and improves more slowly. 
%Both datasets show a critical BPP threshold (>1) where added bits markedly boost reconstruction quality, aligning with elevated progressive ratios. 
In both datasets, we observe a critical threshold near 1 BPP beyond which additional bits yield markedly better reconstruction quality, consistent with higher progressive gains. 
%
% These disparities in initial values and rates arise from inherent point cloud traits: 
These differences reflect intrinsic point-cloud characteristics: SemanticKITTI's LiDAR-derived regular structure and lower complexity enable efficient coding and superior quality at low bitrates (\textit{e.g.}, 0--1 BPP), whereas ShapeNet's diverse, intricate objects require more bits to achieve comparable quality, particularly evident in the low-BPP regime, where SemanticKITTI achieves robust reconstruction with fewer bits.

\subsection{Visualization}
\label{visi}
% In this section, we present visual reconstructions of point cloud models from the SemanticKITTI~\cite{behley2019semantickitti} and ShapeNet~\cite{wu20153d} datasets. 
Figs.~\ref{fig:semantickitti_compression_viz} and~\ref{fig:ShapeNet_compression_viz} depict progressive reconstruction of point clouds from SemanticKITTI and ShapeNet datasets when $\lambda = 0.001$. For instance, in Fig.~\ref{fig:semantickitti_compression_viz}, at a progressive ratio (PR) of 0.03, the essential scene structure remains discernible, preserving core geometry with a minimal bitstream. 
As PR increases, %the decoder incrementally adds channels based on importance ranking, ordered from highest to lowest geometric significance. 
the decoder incrementally activates additional latent channels according to our density-aware importance ranking (from highest to lowest geometric significance). 
This prioritization yields rapid gains in reconstruction quality, highlighting the effectiveness of our density-aware channel importance assessment in favoring %critical geometric and feature data 
geometrically critical coordinates and features for progressive transmission. Notably, Fig.~\ref{fig:ShapeNet_compression_viz} features two objects selected for their sparse-to-dense point distributions. 
%Additionally, reconstructions show minor spatial overlap and noise at low PR.
At very low PR, reconstructions may exhibit minor spatial overlap and noise; %We hypothesize that this arises because, as PR increases and additional bitstreams are decoded, the latent feature information enhances point accuracy while simultaneously reducing erroneous and overlapping points, 
we hypothesize that as PR increases and additional components of the bitstream are decoded, the enriched latent features improve point localization while suppressing spurious and overlapping points, 
thereby improving overall reconstruction performance.

\begin{figure*}[t]
    \centering

    \makebox[0.18\textwidth][c]{\small Ground Truth}
    \hspace{0.05in}
    \makebox[0.18\textwidth][c]{\small PR=0.03}
    \hspace{0.05in}
    \makebox[0.18\textwidth][c]{\small PR=0.09}
    \hspace{0.05in}
    \makebox[0.18\textwidth][c]{\small PR=0.18}
    \hspace{0.05in}
    \makebox[0.18\textwidth][c]{\small PR=0.36}
    % \vspace{0.2cm} 
    
    \includegraphics[width=0.18\textwidth]{figures/vis/model0_org.png}
    \includegraphics[width=0.18\textwidth]{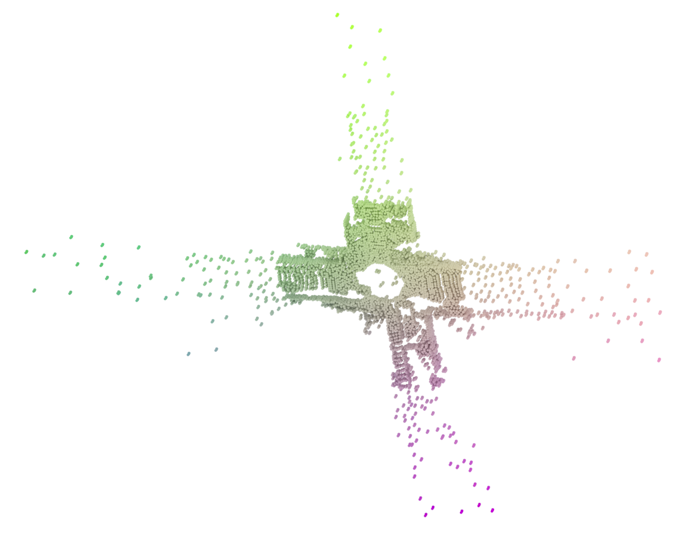}
    \hspace{0.05in}
    \includegraphics[width=0.18\textwidth]{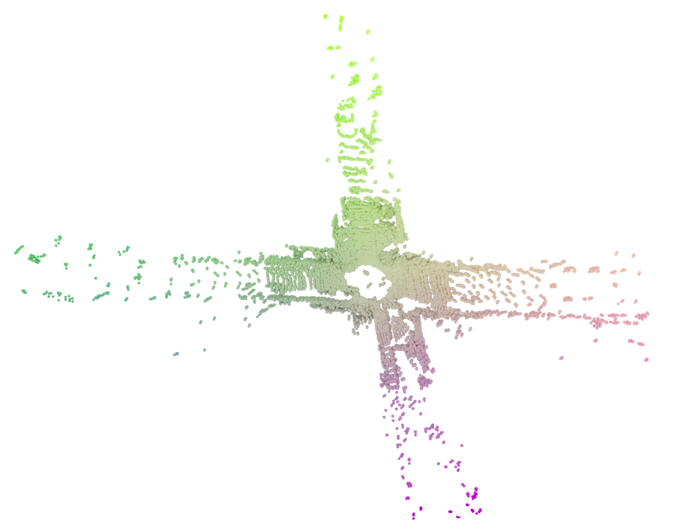}
    \hspace{0.05in}
    \includegraphics[width=0.18\textwidth]{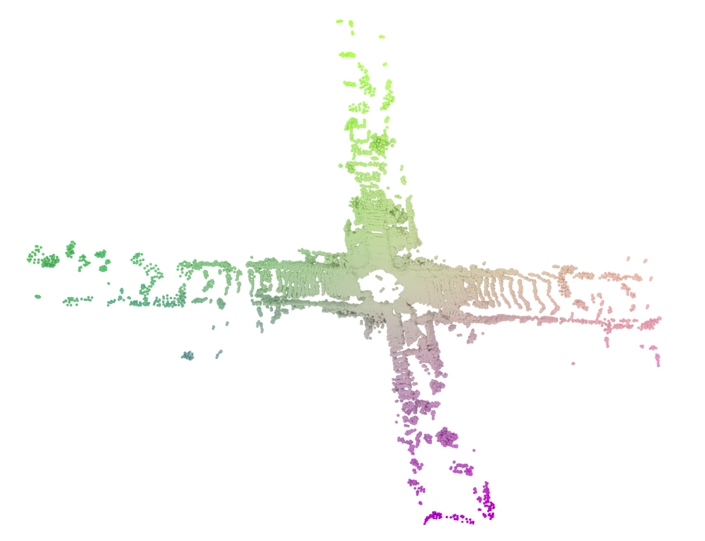}
    \hspace{0.05in}
    \includegraphics[width=0.18\textwidth]{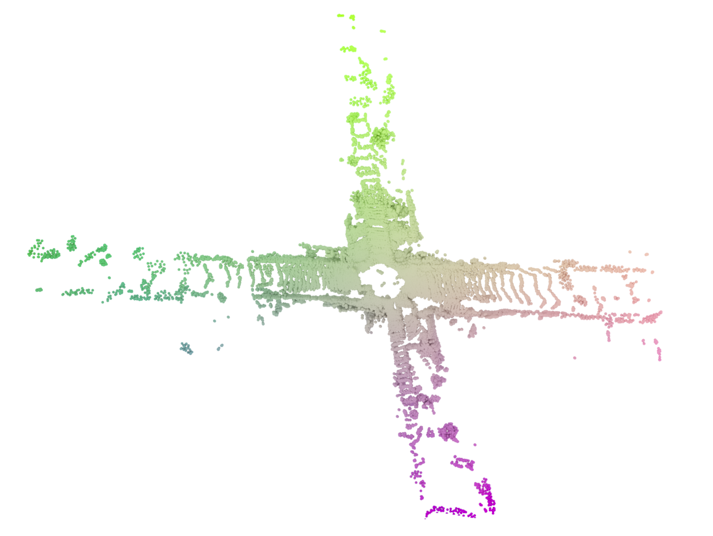}
    
    % \vspace{0.2cm}
    
    \includegraphics[width=0.173\textwidth]{figures/vis/model1_org.png}
    \hspace{0.05in}
    \includegraphics[width=0.173\textwidth]{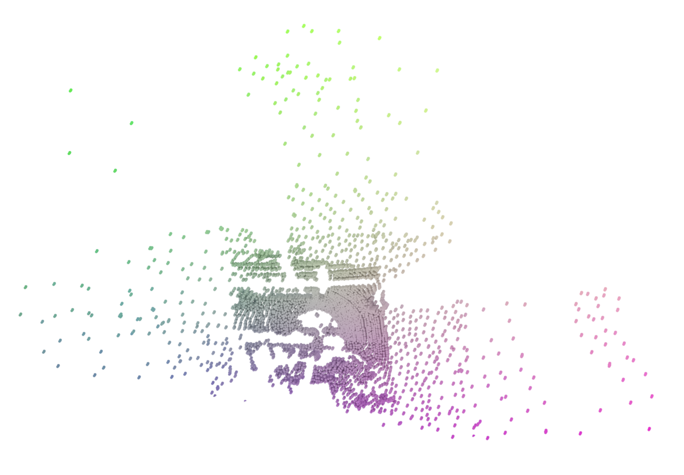}
    \hspace{0.05in}
    \includegraphics[width=0.173\textwidth]{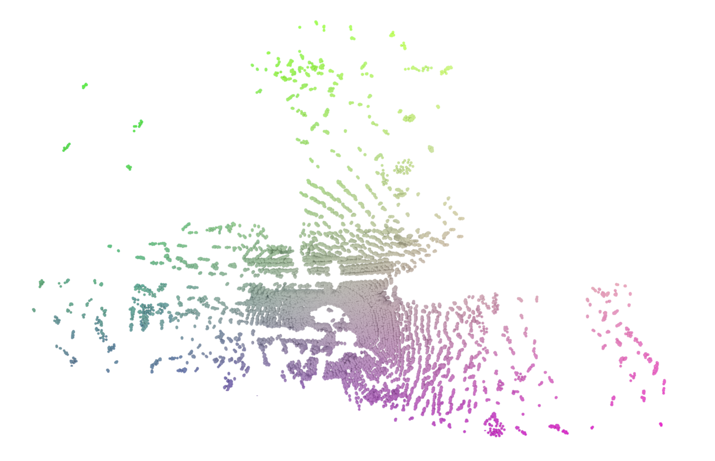}
    \hspace{0.05in}
    \includegraphics[width=0.173\textwidth]{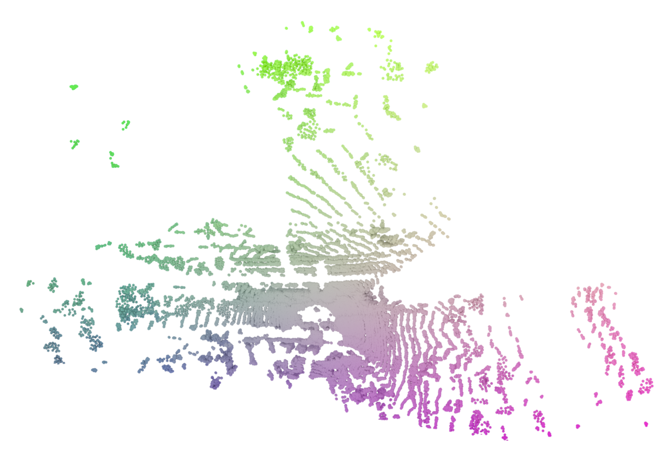}
    \hspace{0.05in}
    \includegraphics[width=0.173\textwidth]{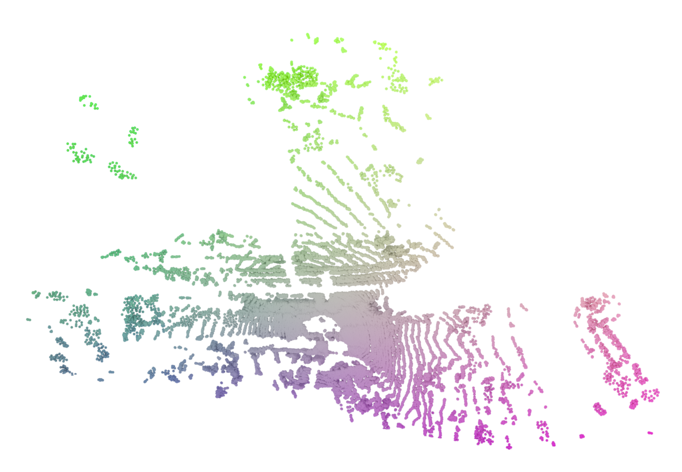}
    % \vspace{0.2cm} % Vertical space after the row

    \caption{%Visualization of the progressive compression objective applied to three distinct models from the SemanticKITTI dataset, illustrating the impact of feature dropping only. 
    Visualization of progressive coding with feature-only drop applied to two SemanticKITTI models. 
    From left to right: Ground Truth, PR = 0.03, 0.09, 0.18, and 0.36, yielding the corresponding drop rates (1-PR) of 0.97, 0.91, 0.82, and 0.64, respectively.}
    \label{fig:semantickitti_compression_viz_Feature} % Added a label for easier referencing
    \label{fig9}
    \vspace{-1em}
\end{figure*}

\subsection{Ablation Studies}

\label{ablation_dropping_strategies}
\begin{figure}[!t]
    \centering
    % \subfloat[BPP vs PR for Different Drop Strategies]
    \subfloat[]{\includegraphics[width=0.23\textwidth]{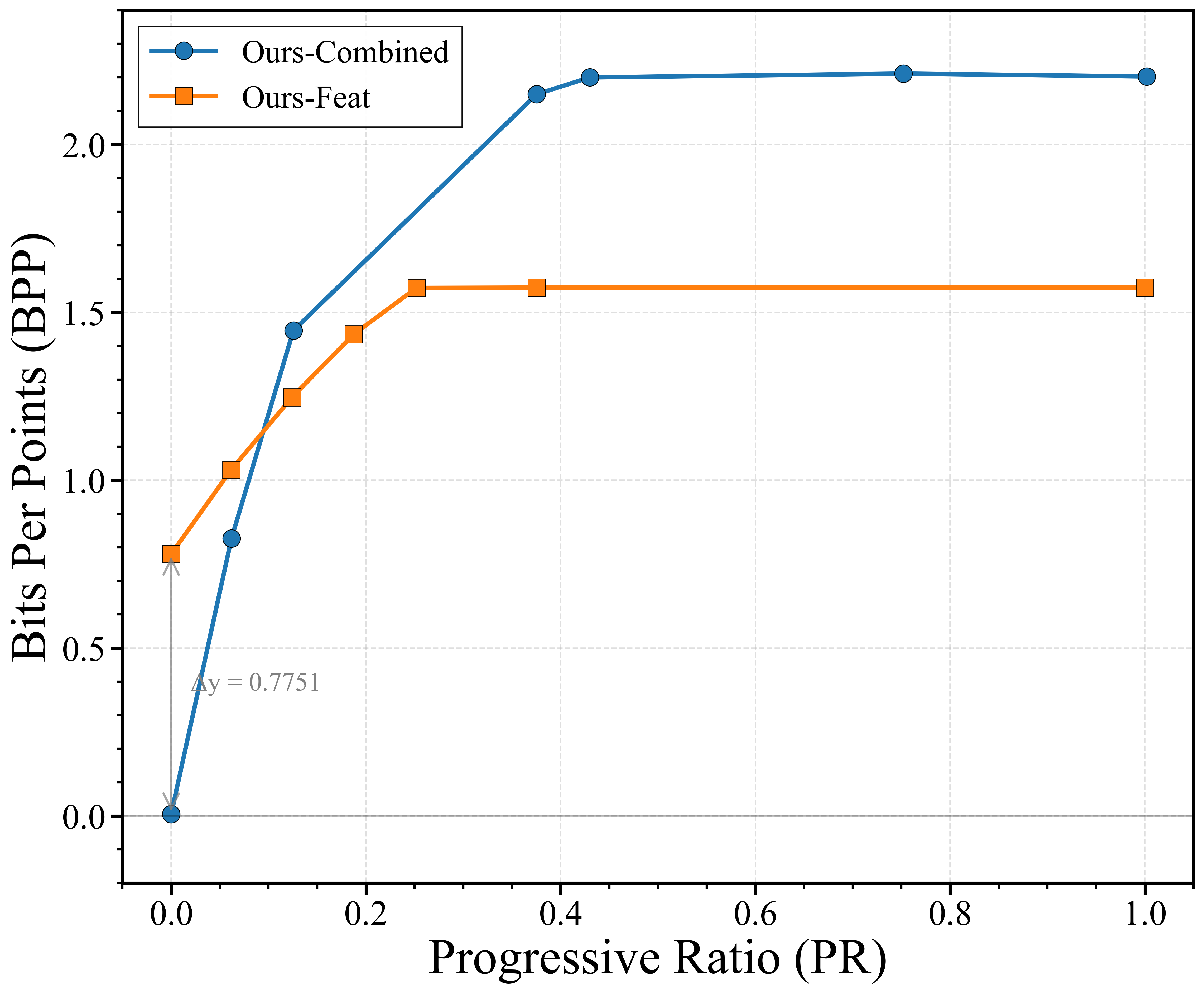}
    \label{subfig:semantickitti_progressive_both_bpp_compare}}
    \hfill
    % \subfloat[PSNR-D2 vs BPP for Feature-Only Drop with varying $\lambda$ values]
    \subfloat[]{\includegraphics[width=0.23\textwidth]{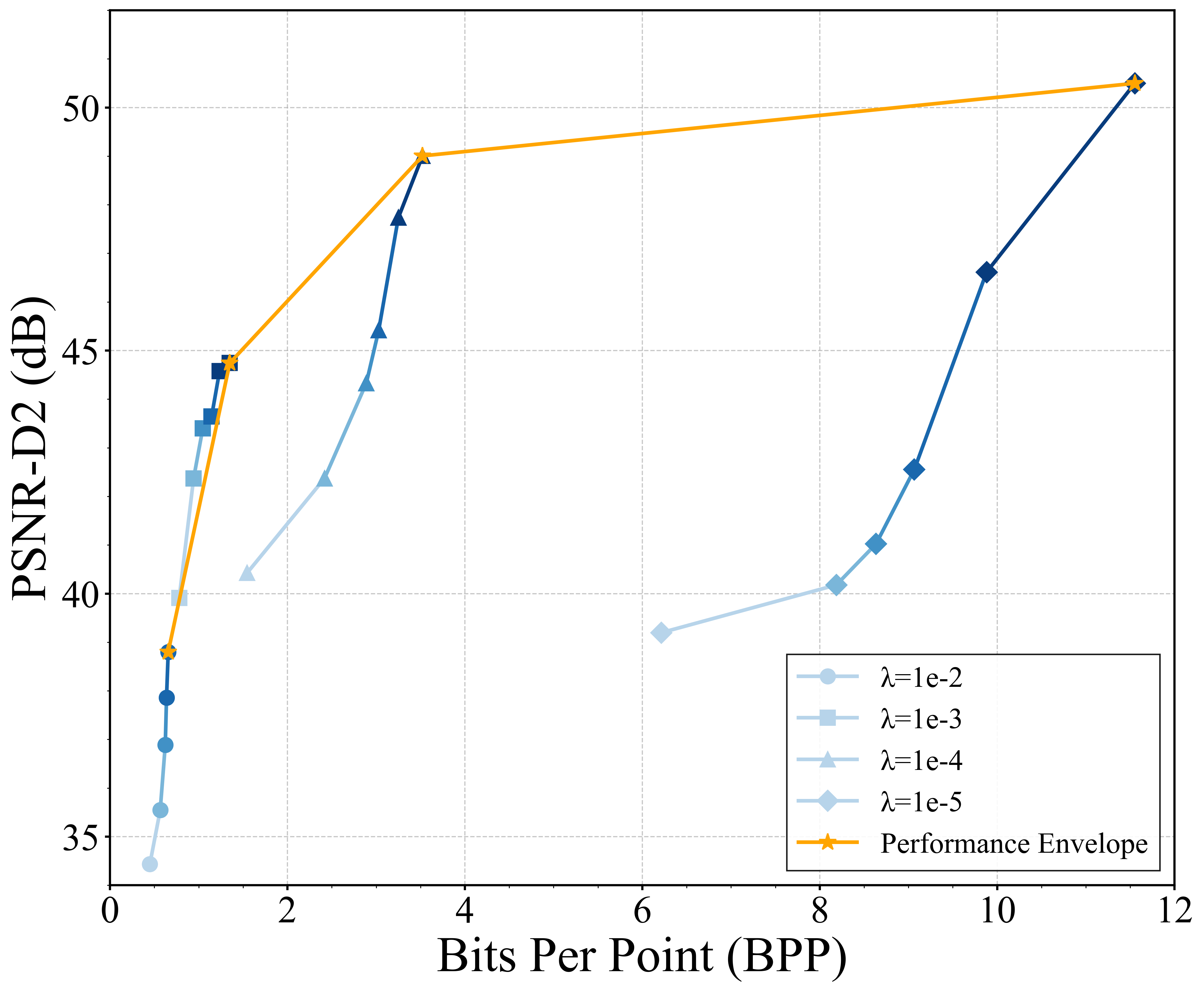}
    \label{subfig:semantickitti_progressive_feature}}
    \caption{Ablation study of progressive coding for different drop strategies and $\lambda$ values on SemanticKITTI: (a) BPP vs PR for different drop strategies, and (b) PSNR-D2 vs BPP for feature-only drop with varying $\lambda$ values.}
    \label{fig:quantitative_results_featre_only_drops}
    \vspace{-1em}
\end{figure}

%Two strategies were investigated: 
We compare two strategies: combined coordinate-feature drop and feature-only drop, where models were trained with %$\lambda$ values ranging from $10^{-2}$ to $10^{-5}$. 
$\lambda \in [10^{-2}, 10^{-5}]$. 
Each model was trained once per $\lambda$ and subsequently evaluated through progressive testing across multiple progressive ratios. Results are displayed in Fig.~\ref{fig:quantitative_results_1}, Fig.~\ref{subfig:semantickitti_progressive_both}, and Fig.~\ref{subfig:semantickitti_progressive_feature}, which indicate the higher PSNR-D2 performance for the combined-drop method, underscoring its effectiveness through BD-Rate improvements exceeding 12.3\% on the SemanticKITTI dataset and and 9.6\% on ShapeNet dataset compared to the feature-only strategy statistically.

% \begin{figure}[!ht]
%     \centering
%     % \subfloat[BPP vs PR for Different Drop Strategies]
%     \subfloat[]{\includegraphics[width=0.24\textwidth]{figures/result/comparison_plot.png}\label{subfig:semantickitti_progressive_both_bpp_compare}}
%     \hfill
%     % \subfloat[PSNR-D2 vs BPP for Feature-Only Drop with varying $\lambda$ values]
%     \subfloat[]{\includegraphics[width=0.24\textwidth]{figures/result/BPP_vs_psnr_progressive_feature.png}\label{subfig:semantickitti_progressive_feature}}
%     \caption{Ablation study of progressive coding for different drop strategies and $\lambda$ values on SemanticKITTI: (a) BPP vs PR for different drop strategies, and (b) PSNR-D2 vs BPP for feature-only drop with varying $\lambda$ values.}
%     \label{fig:quantitative_results_featre_only_drops}
% \end{figure}

% Fig.~\ref{subfig:semantickitti_progressive_both_bpp_compare} shows that, for $\lambda = 0.001$, the combined drop strategy spans a wider BPP range (0.008--2.2) than the feature-only approach (0.77--1.57). However, the feature-only strategy reaches its maximum BPP at a lower progressive ratio (PR = 0.25 vs. 0.43 for combined), highlighting its greater efficiency in utilizing progressive data.
%Furthermore, comparison between Figs.~\ref{subfig:semantickitti_progressive_both} and~\ref{fig:quantitative_results_featre_only_drops} reveals distinct reconstruction behavior, where the performance envelopes for different strategies are color-coded in red and orange colors. % to differentiate the performance envelopes for different strategies. 

Besides, we set $\lambda = 0.001$ here to show the visualizations of different drop strategies' results and get Fig.~\ref{fig:semantickitti_compression_viz_Feature}, which is consistent with the setting of Sect.~\ref{visi}. Compared with Fig.~\ref{fig:semantickitti_compression_viz}, even though feature-only drop shows strong central and structural information when PR is low (\textit{e.g.}, 0.03), its BPP starts from a much higher value, which further proves the benefit of combined-drop that it can achieve the progressive coding starting from a much smaller BPP value. In details, the feature-only method shows an elevated initial BPP of 0.77 because of direct coordinate encoding, in contrast to the combined-drop approach starting at 0.008 BPP, which can be observed from BPP-PR relationships Fig.~\ref{subfig:semantickitti_progressive_both_bpp_compare}.

\section{Conclusion}
\label {sect:conclusion}

In this work, we presented ProDAT, a novel framework for progressive point cloud coding that introduces Tail-drop, an innovative data selection method that strategically discards less critical information while maintaining reconstruction quality. 
%Our approach represents the first successful demonstration of Tail-drop's effectiveness in progressive coding, establishing a new paradigm for efficient point cloud processing.
Unlike conventional methods, ProDAT achieves progressive coding with just a single training iteration, significantly reducing computational overhead while maintaining competitive performance.
%We validated our framework on mainstream datasets (SemanticKITTI and ShapeNet), demonstrating superior BD-Rate performance compared to state-of-the-art (SOTA) methods.
Comprehensive experiments conducted on %two widely adopted 
SemanticKITTI and ShapeNet datasets demonstrate superior BD-Rate performance compared to state-of-the-art (SOTA) methods.
Our exploration of density-aware Tail-drop and latent space drop strategies provides new insights into coding dynamics, revealing opportunities for further optimization.
% Looking ahead, future research could enhance ProDAT by developing more sophisticated criteria for identifying and preserving critical latent or structural information and investigating adaptive drop strategies that optimize coding in real time based on scene complexity.

\bibliography{references}

\end{document}